\documentclass[10pt,twocolumn,letterpaper]{article}

\usepackage[pagenumbers]{cvpr} 

\usepackage{algorithm}
\usepackage{algorithmic}
\usepackage{multirow}
\usepackage{tabularx}
\usepackage{amsfonts,bm}
\usepackage{threeparttable}
\usepackage{upgreek}
\usepackage{pifont}%
\usepackage{bbding}
\usepackage{makecell}
\usepackage{color}
\usepackage{xcolor}
\usepackage{colortbl}
\usepackage{wrapfig}
\usepackage{soul}
\usepackage{wasysym}
\usepackage{xspace}
\usepackage{subcaption}
\usepackage[first=0,last=9]{lcg}
\definecolor{Gray}{gray}{0.85}

\newcolumntype{x}[1]{>{\centering\arraybackslash}p{#1pt}}
\newcolumntype{y}[1]{>{\raggedright\arraybackslash}p{#1pt}}
\newcolumntype{z}[1]{>{\raggedleft\arraybackslash}p{#1pt}}

\definecolor{cvprblue}{rgb}{0.21,0.49,0.74}
\usepackage[pagebackref,breaklinks,colorlinks,citecolor=cvprblue]{hyperref}

\def\paperID{2812} 
\def\confName{CVPR}
\def\confYear{2024}

\title{\includegraphics[width=0.4cm,height=0.4cm]{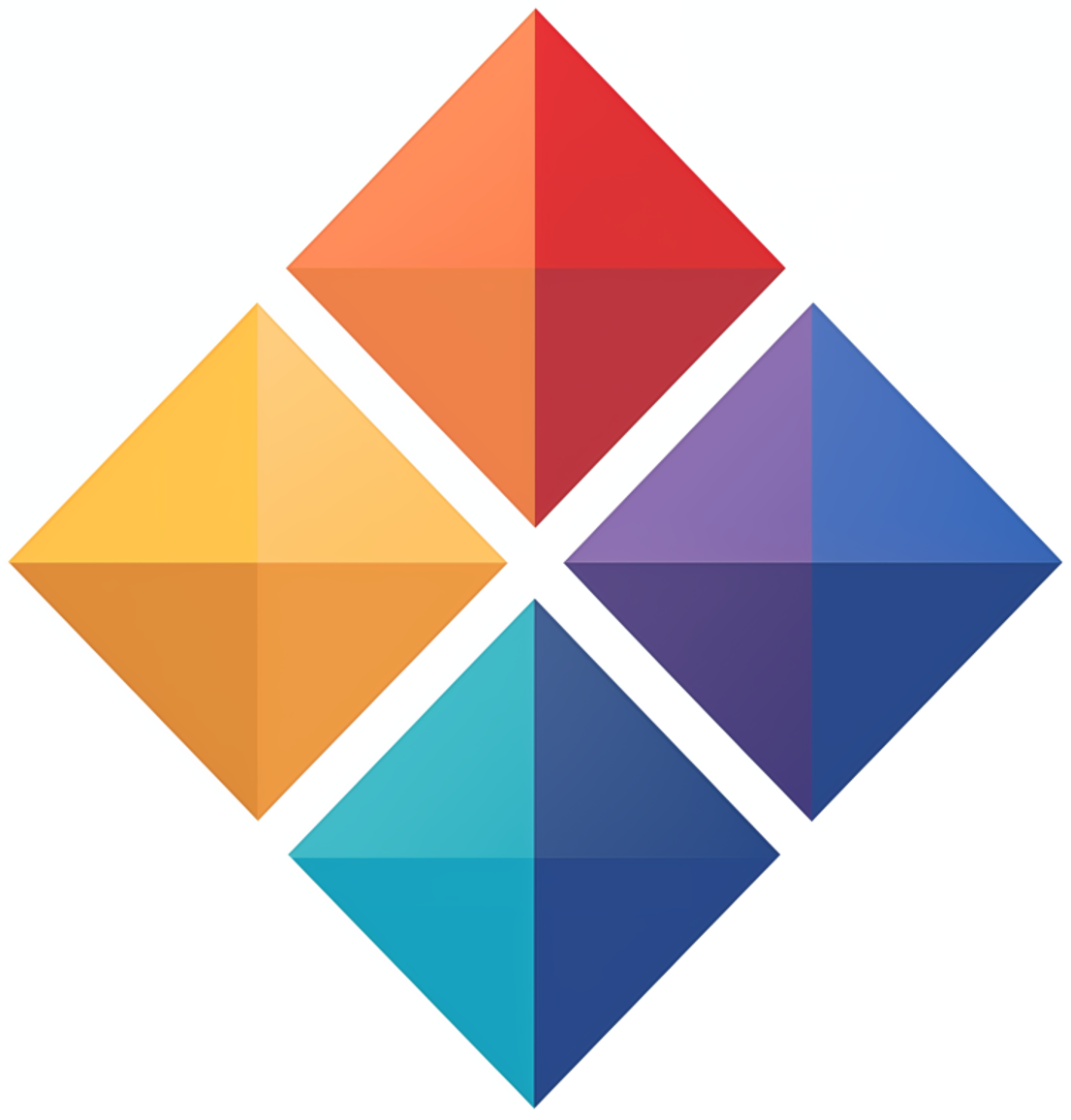} ChatterBox: Multi-round Multimodal Referring and Grounding}

\author{
Yunjie Tian\footnotemark[1]\\
UCAS\\
{\tt\small tianyunjie19@mails.ucas.ac.cn}
\and
Tianren Ma\footnotemark[1]\\
UCAS\\
{\tt\small matianren18@mails.ucas.ac.cn}
\and
Lingxi Xie\\
Huawei Inc.\\
{\tt\small 198808xc@gmail.com}
\and
Jihao Qiu\\
UCAS\\
{\tt\small qiujihao19@mails.ucas.ac.cn}
\and
Xi Tang\\
UCAS\\
{\tt\small tangxi19@mails.ucas.ac.cn}
\and
Yuan Zhang\\
UCAS\\
{\tt\small zhangyuan192@mails.ucas.ac.cn}
\and
Jianbin Jiao\\
UCAS\\
{\tt\small jiaojb@ucas.ac.cn}
\and
Qi Tian\\
Huawei Inc.\\
{\tt\small tianqi1@huawei.com}
\and
Qixiang Ye\\
UCAS\\
{\tt\small qxye@ucas.ac.cn}
}

\newcommand{\task}{MRG\xspace}
\newcommand{\benchmark}{CB-300K\xspace}
\newcommand{\model}{ChatterBox\xspace}
\newcommand{\dataA}{CB-MRG\xspace}
\newcommand{\dataB}{CB-LC\xspace}
\newcommand{\dataC}{CB-REF\xspace}
\newcommand{\dataD}{CB-GND\xspace}

\begin{document}
\twocolumn[{
\renewcommand\twocolumn[1][]{#1}
\maketitle
\vspace{-20pt}
\begin{center}
\centering
\includegraphics[width=1.0\linewidth]{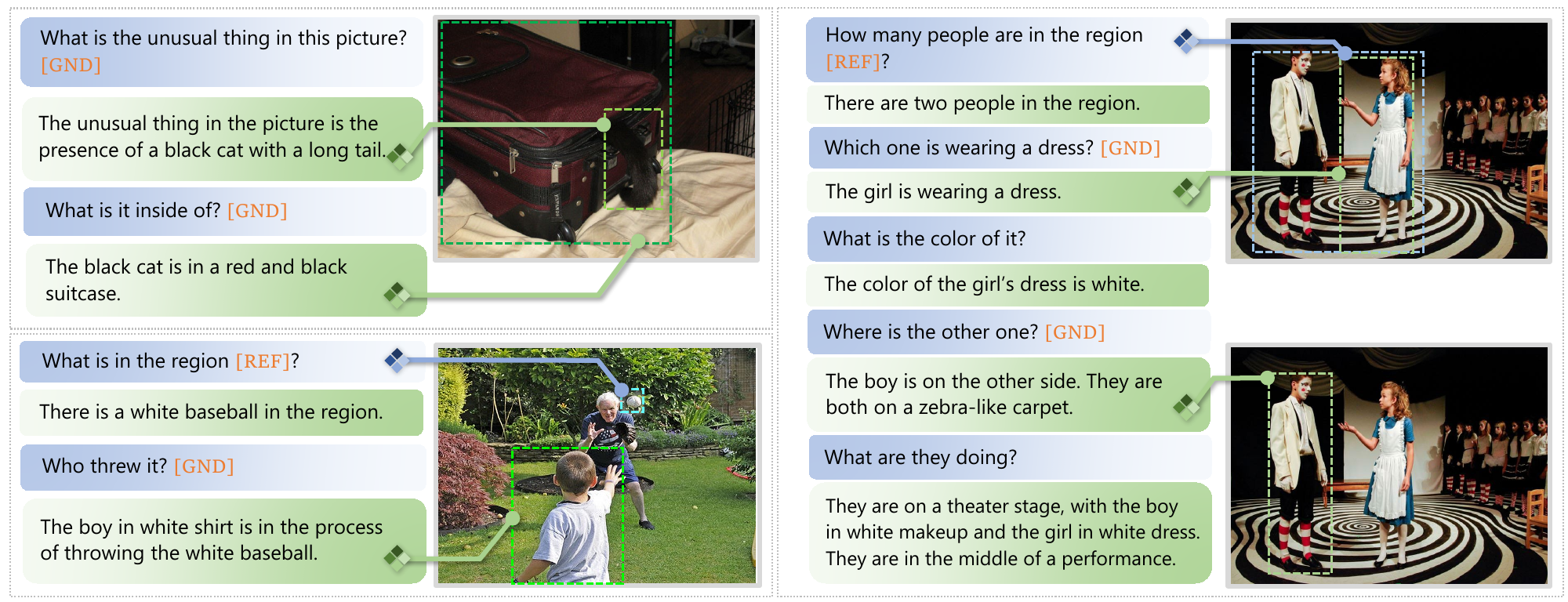}
\vspace{-20pt}
\captionof{figure}{A showcase of the multi-round referring and grounding (\task) task. During the dialogue, the agent can receive either a \texttt{[REF]} token for referring expressions or a \texttt{[GND]} token for visual grounding; without these tokens, the task becomes generic visual question answering. All the answers are generated by the \model agent, demonstrating its strong ability in visual recognition. In particular, \model can understand \textbf{logically related} questions and incorporate contextual information to provide answers. For instance, in the right-hand thread, the question `Where is the other one?' necessitates the agent to recognize that `one' refers to a person and then locate the `other' person distinct from the one mentioned earlier.}
\label{fig:task}
\end{center}
\vspace{10pt}
}]

\renewcommand{\thefootnote}{*}
\footnotetext[1]{Equal contribution.}

\begin{abstract}

In this study, we establish a baseline for a new task named multimodal multi-round referring and grounding (\task), opening up a promising direction for instance-level multimodal dialogues. We present a new benchmark and an efficient vision-language model for this purpose.
The new benchmark, named \benchmark, spans challenges including multi-round dialogue, complex spatial relationships among multiple instances, and consistent reasoning, which are beyond those shown in existing benchmarks. 
The proposed model, named \model, utilizes a two-branch architecture to collaboratively handle vision and language tasks. 
By tokenizing instance regions, the language branch acquires the ability to perceive referential information. Meanwhile, ChatterBox feeds a query embedding in the vision branch to a token receiver for visual grounding. 
A two-stage optimization strategy is devised, making use of both \benchmark and auxiliary external data to improve the model's stability and capacity for instance-level understanding.
Experiments show that \model outperforms existing models in \task both quantitatively and qualitatively, paving a new path towards multimodal dialogue scenarios with complicated and precise interactions. 
Code, data, and model are available at: \textit{\textcolor{red}{https://github.com/sunsmarterjie/ChatterBox}}.

\end{abstract}

\section{Introduction}
\label{sec:intro}

Large language models (LLMs) have shown impressive capabilities across a wide range of natural language tasks~\cite{brown2020language}. 
In the computer vision community, researchers have integrated LLMs with images and videos, creating a series of multimodal large language models (MLLMs)~\cite{alayrac2022flamingo,liBLIP2BootstrappingLanguageImage2023,liuVisualInstructionTuning2023,li2022blip}.
Recently, with the success of instruction tuning, comprehensive image-text instruction datasets have been constructed~\cite{zhangGPT4RoIInstructionTuning2023,liuVisualInstructionTuning2023,wang2023cogvlm, laiLISAReasoningSegmentation2023}, empowering the ability of MLLMs in a wide range of multimodal understanding tasks. 

We argue that a powerful multimodal agent should have the ability to understand logically related questions and perform basic vision-aware tasks such as referring and grounding (see Figure~\ref{fig:task}). However, few existing models were equipped with all these abilities. To fill up the empty, we propose the multi-round multimodal referring and grounding (\task) task, where the MLLM is expected to engage in referring (\textit{i.e.}, recognizing a designated region or object) and grounding (\textit{i.e.}, locating a region or object from the image) tasks at any time in a multi-round dialogue, meanwhile retaining the consistent logic of the entire conversation. \task is similar to the behavior in which humans interact with agents.

We make two-fold contributions to enable \task. Firstly, we establish a new benchmark named \benchmark, which comprises the first-ever image-text dataset for \task and an evaluation metric that takes the accuracy of both visual and linguistic understanding into consideration. The construction of \benchmark mainly builds upon Visual Genome~\cite{krishna2017visual}, where we feed the metadata into GPT-4~\cite{openaiGPT4TechnicalReport2023} and prompt it to generate multi-round dialogues with referring and grounding requests. Post-processing is then performed to guarantee the correctness of dialogues and organize them to subsets for various purposes.

Secondly, we propose an MLLM named \model to solve the challenging task. The key lies in the integration of visual and linguistic information. To fulfill this purpose, we design a two-branch architecture, where the language branch understands the logic of the question, and the vision branch plays the role of visual feature extraction and recognition (\textit{e.g.}, grounding). The \model design can be easily understood and optimized, requiring only 15 GPU-days for a two-stage optimization process that involves both \benchmark and auxiliary external data (\textit{e.g.}, RefCOCO~\cite{kazemzadeh2014refcoco} and LLaVA-Instruction-150K~\cite{liuVisualInstructionTuning2023}).

We conduct both quantitative and qualitative studies on the \benchmark benchmark and validate \model's superiority over existing models in \task. Some examples of \model performing \task are displayed in Figure~\ref{fig:task}. \model also transfers to easier tasks (\textit{e.g.}, single-round dialogue, referring, grounding) seamlessly. Our research advocates that delicate and precise interactions are strongly required to enhance the ability of multimodal dialogue as well as artificial general intelligence systems.

\newcommand{\tick}{\textcolor{green}{\CheckmarkBold}\xspace}
\newcommand{\cross}{\textcolor{red}{\XSolidBrush}\xspace}
\begin{table}[!t]
\setlength{\tabcolsep}{0.08cm}
\small
\centering
\caption{A comparison of \model to recent studies \textit{w.r.t.} the abilities to perform multi-round dialogues (including region-level referring and visual grounding), the proposal of new data ($\dagger$: it involves generating new dialogue data rather than simply reorganizing existing data), and training costs. N/R: not reported.}
\vspace{-4pt}
\begin{tabular}{l|c|c|c|c|c}
\toprule
\multirow{2}{*}{Method} & Multi- & Region & Visual & New & Training \\
& Round & Referring & Grounding & Data$^\dagger$ & (G-days) \\
\midrule
%
LLaVA~\cite{liuVisualInstructionTuning2023}        & \tick  & \cross & \cross & \tick  & 4.7  \\
InstructBLIP~\cite{instructblip}                   & \tick  & \cross & \cross  & \cross  & 24  \\
VisionLLM~\cite{wang2023visionllm}                 & \cross & \tick  & \tick  & \cross & N/R \\
Kosmos-2~\cite{pengKosmos2GroundingMultimodal2023} & \tick  & \tick  & \tick  & \tick  & 256 \\
GPT4RoI~\cite{zhangGPT4RoIInstructionTuning2023}   & \tick  & \tick  & \cross & \cross & N/R \\
LISA~\cite{laiLISAReasoningSegmentation2023}       & \tick  & \cross & \tick & \cross & 8   \\
\bottomrule
\rowcolor{Gray}
\textbf{ChatterBox }                         & \tick  & \tick  & \tick  & \tick  & 15 \\
\bottomrule
\end{tabular}

\label{tab:comparison}
\end{table}

We compare our work to recent studies on multimodal dialogue in Table~\ref{tab:comparison} and summarize our contributions below:
\begin{itemize}
\item We introduce a new task named multi-round multimodal referring and grounding (\task).
\item We propose a data construction scheme and establish the \benchmark benchmark to facilitate the research in \task.
\item We present \model, a vision-language model that injects explicit vision modules into an MLLM, providing an agile and effective solution of \task.
\end{itemize}

\begin{figure*}[!t]
\centering
\includegraphics[width=1\textwidth]{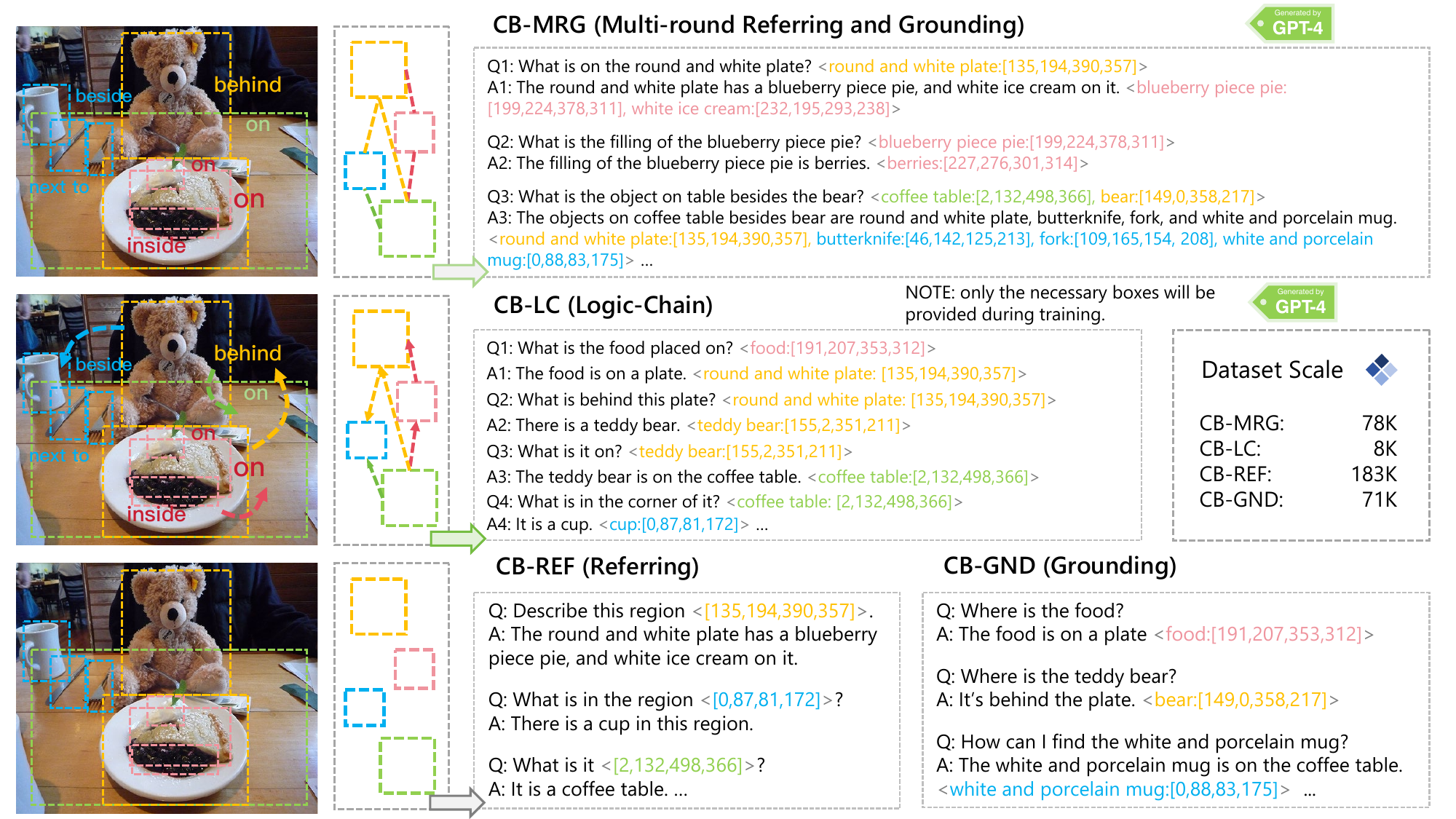}
\caption{The \benchmark data contains four subsets for different purposes. The images and metadata (object locations and descriptions) are inherited from Visual Genome. The same image can appear in different subsets. The former two subsets, \dataA and \dataB, are obtained by prompting GPT-4 to read the metadata and generate questions and answers. The latter two subsets, \dataC and \dataD, are produced using manually designed rules and then polished by GPT-3.5. \textit{This figure is best viewed in color.}}
\label{fig:dataset_examples}
\end{figure*}

\section{Related Work}
\label{relatedwork}

\subsection{Multimodal Large Language Models}
Large language models~\cite{devlin2018bert,touvron2023llama, brown2020language,chiang2023vicuna,chung2022scaling, zeng2022glm, thoppilan2022lamda, chowdhery2022palm, zhang2022opt} have opened a new era of AI, demonstrating the potential to create a generalist model that can even cover different modalities. The computer vision community has witnessed a trend of unifying vision and language data using multimodal large language models~\cite{li2022blip, liBLIP2BootstrappingLanguageImage2023, alayrac2022flamingo, liuVisualInstructionTuning2023}. The pioneering efforts involved projecting vision and language data into the same feature space~\cite{radford2021learning,alayrac2022flamingo}. Attempts to adapt a LLM to visual tasks have been made internally or externally: Flamingo~\cite{alayrac2022flamingo} interleaves cross-attention blocks within a LLM for visual-language alignment. BLIP-2~\cite{liBLIP2BootstrappingLanguageImage2023} proposes Q-former, an external block that uses multiple vision-language losses to align queried visual features with text.


\subsection{Multimodal Instruction Data}
Later, inspired by the instruction tuning mechanism~\cite{ouyang2022training} of the GPT series, MLLMs started collecting instruction data from various sources. One of the early efforts was visual instruction tuning~\cite{liuVisualInstructionTuning2023} which provides a novel method for data construction. By feeding external metadata including bounding box annotations and textual descriptions into GPT-4 with designed prompts, comprehensive detailed conversations about the images can be generated without vision access. The idea was followed by other works~\cite{chenPositionEnhancedVisualInstruction2023,chenShikraUnleashingMultimodal2023} to harvest various types of instruction data. In another approach, the image feature was fed into an MLLM and prompted for instruction data~\cite{zhu2023minigpt}. Additionally, richer information (\textit{e.g.}, phrase grounding) was also collected~\cite{pengKosmos2GroundingMultimodal2023} with the assistance of external vision-language models, such as GLIP~\cite{liGroundedLanguageImagePretraining2022a}). The new data and learning strategy enabled more abilities to emerge via multimodal dialogue~\cite{liuVisualInstructionTuning2023,gong2023multimodal,alayrac2022flamingo,wang2023cogvlm,youFerretReferGround2023,laiLISAReasoningSegmentation2023}.

\subsection{Instance-Level Understanding of MLLMs}
MLLMs can be largely enhanced by the ability of instance-level understanding, \textit{i.e.}, the models can (1) respond to questions targeted at specified regions of the image and (2) find regions that correspond to the contents in the dialogue. We address these two abilities as visual referring~\cite{zhangGPT4RoIInstructionTuning2023,chenPositionEnhancedVisualInstruction2023} and visual grounding~\cite{pengKosmos2GroundingMultimodal2023,liu2023grounding}, respectively, and they have been well studied in the computer vision community under various task designations and settings. Integrating them into MLLMs, however, remains a challenge. There are two main approaches to integration, differing in whether to encode the position information explicitly or not. Explicit methods~\cite{pengKosmos2GroundingMultimodal2023,wang2023visionllm} are easier to  optimize and explain by introducing location tokens, while implicit methods~\cite{chenShikraUnleashingMultimodal2023,wang2023cogvlm,xuanPinkUnveilingPower2023} offer greater flexibility. Nonetheless, these multimodal methods are not well-suited for handling instance-level multi-round dialogues, which can lead to superficial interaction.

\section{The \benchmark Benchmark}
\label{benchmark}

We establish a benchmark named ChatterBox-300K with the aim to enhance the ability of multimodal dialogue systems in multi-round referring and grounding. Similar to previous work~\cite{liuVisualInstructionTuning2023}, the dataset is mainly constructed upon context-rich image data (where we use the Visual Genome dataset~\cite{krishna2017visual} for the richness in the annotation of object-level relationship) and assisted by a large language model (where we use GPT-4 as an off-the-shelf model for understanding). We also define a metric for evaluating the models in this new scenario.

\subsection{Data Collection}
\label{benchmark:data}

When an image is sampled from Visual Genome, we refer to the annotation data which mainly has three parts: (1) objects with bounding boxes (\textit{e.g.}, there is a man at $[x_1,y_1,w_1,h_1]$ and a computer at $[x_2,y_2,w_2,h_2]$), (2) the relationship between objects (\textit{e.g.}, the man is operating the computer), (3) auxiliary attributes of objects (\textit{e.g.}, the man is in black). We summarize all the information (in pure texts) as contexts, feed them to GPT-4, and ask it to generate question-and-answer pairs of different aspects. An additional request for GPT-4 is that, in each sequence of  consecutive question-and-answer pairs, the latter questions should build on the former ones, so that we can fully test the model's ability to engage in multi-round dialogue.

In summary, there are four subsets in \benchmark including a generic subset for \task and three specifically designed subsets, as illustrated in Figure~\ref{fig:dataset_examples}. We elaborate the design principles as follows and leave the technical details (\textit{e.g.}, the full prompts for GPT-4, and the Python code for data generation and filtering) in the supplementary material.
\begin{itemize}
\item\textbf{\dataA} is a generic data collection for \textbf{\task}. We ask GPT-4 to generate questions and answers that focus on instance-level relationship. Provided with related instances, GPT-4 shall write dialogues using their relationship. Meanwhile, their location (as a bounding box, formed $[x,y,w,h]$) are also attached in each question and answer, retaining the maximum spatial information for further referring and grounding tasks. We also write additional prompts, asking GPT-4 to create multi-round referring and grounding requests, which is achieved by asking about further relationships upon the instance(s) that appeared in the former answers.
\item\textbf{\dataB} is a subset that extends the ability for \textbf{logic-chain} \task. The key differences include (1) adding strict restrictions upon the prompt we used to generate \dataA (\textit{e.g.}, each question must be built upon exactly one aforementioned relationship), (2) deleting invalid question-and-answer pairs using manually-designed rules (\textit{e.g.}, logic-deficient dialogues, mismatched or missing boxes, \textit{etc.}; more details are provided in the supplementary material), and (3) calling GPT-4 again to check the entire thread, cleaning up incorrect descriptions and contradictions. The strict filtering procedure makes \dataB a high-quality subset for logic-chain \task, but the size is relatively small\footnote{This is partly caused by the limited ability of GPT-4 in understanding the prompt and generating correct question-and-answer pairs (it is a probabilistic LLM that can make errors). We expect that more logic-chain data can be generated in the future with stronger LLMs as the AI assistant.}.
\item\textbf{\dataC} adds more dialogues for \textbf{referring expression}. These dialogues are generated by a manually designed rule that extracts the annotated bounding box(es) as referential inputs and the description of the region as answers. Various Q\&A styles are applied and GPT-3.5 is used to polish the grammatical issues.
\item\textbf{\dataD} adds more dialogues for \textbf{visual grounding}. These dialogues are generated manually using the reversed rules of \dataC.
\end{itemize}

During the generation procedure, the description and bounding boxes are available for each instance in each question. To facilitate multi-round referring, we post-process the data, using the pronoun `it' to replace the whole unit. The processing details are different between training and testing and will be elaborated in the following sections.

Table~\ref{tab:statistics} displays the statistics of \benchmark. We randomly split the dataset by leaving out $800$ and $200$ threads from \dataA and \dataB for testing and using the remaining data for training. We guarantee that different occurrences of the same image (in different subsets) do not appear in the training and testing splits simultaneously.

\begin{table}[t]
\setlength{\tabcolsep}{0.2cm}
\centering
\caption{The number of threads and the number of question-and-answer pairs of each individual subset and the entire benchmark. }
\vspace{-4pt}
\begin{tabular}{l|r|r}
\toprule
\textbf{Set} & \textbf{\# threads} & \textbf{\# Q\&A pairs} \\
\midrule
\dataA     &  77,814 & 437,229 \\
\dataB     &   7,834 &  25,617 \\
\dataC     & 183,446 & 183,446 \\
\dataD     &  70,783 &  70,783 \\
\midrule
\benchmark & 339,877 & 717,075 \\
\bottomrule
\end{tabular}
\label{tab:statistics}
\end{table}

\noindent\textbf{Comparison to other datasets.}
\label{sec:com}
The \benchmark dataset differs from existing datasets for multimodal dialogues, such as LLaVA-Instruction-150K~\cite{liuVisualInstructionTuning2023}, in the following aspects.
\begin{itemize}
\item \benchmark focuses on recognizing instance-level information, reflecting in a large amount of visual referring and grounding requests. With a bounding box available for each instance, \benchmark offers the ability to localize and describe instances more accurately, which improves the granularity of dialogues.
\item \benchmark constructs a large number of threads for \task, forcing the model to gain the ability to perform visual recognition based on logic chains.
\item \benchmark is a versatile dataset, which implies that it can be used for various purposes. A typical example lies in using different subsets to train individual yet complementary abilities (\textit{e.g.}, referring, grounding, multi-round understanding, \textit{etc.}) and then combining them into a strong \task system.
\end{itemize}

\begin{figure*}[!t]
\centering
\includegraphics[width=\linewidth]{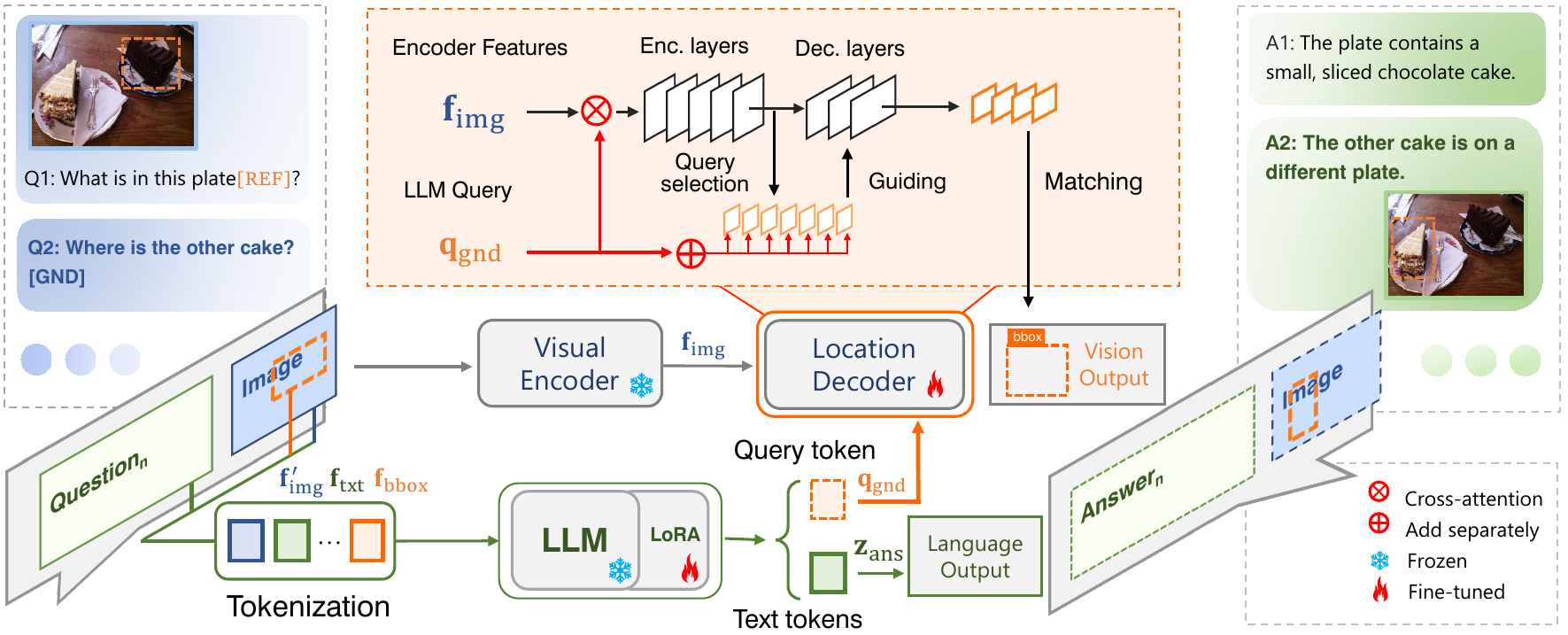}
\caption{The architecture of the \model model. It receives the image and the current question with dialogue history as input, and produces language output and, if necessary, vision output (\textit{i.e.}, visual grounding results). The location decoder is magnified to illustrate the interaction between the query token and visual features. \textit{This figure is best viewed in color.}}
\label{fig:framework}
\end{figure*}

\subsection{Evaluation Metric}
\label{benchmark:metric}

Evaluation is an important issue for multimodal dialogues, particularly for this study which involves logic-chain multifaceted abilities including multi-round reasoning, question answering, and visual grounding. 
We notice that recent studies (\textit{e.g.}, ~\cite{youFerretReferGround2023}) have applied state-of-the-art LLMs for benchmarking, but it incurs two-fold risks: (1) GPT-4's speculative sampling strategy and online update brings unstable randomness for evaluation; meanwhile, (2) GPT-4 developed certain preferences during instruction tuning, making its quantitative decisions biased.

Within a single round, the requirement is similar to the task of grounded image captioning~\cite{zhou2020more, li2023gligen}, where we employ two scores for evaluation. The first term focuses on the language part (\textit{i.e.}, whether the answer is linguistically correct). We refer to the BERT score~\cite{zhang2019bertscore} to compute the similarity between the model's output and the ground-truth answer (we use the RoBERTa-large model~\cite{liu2019roberta}). The second term focuses on the visual grounding part (\textit{i.e.}, whether the detected bounding box is accurate), and the IoU between the detected and ground-truth boxes is naturally taken into consideration.

In summary, if there is no request for grounding (\textit{i.e.}, $M=0$), the single-round score equals $\mathrm{BERT}(\mathbf{a}_m,\mathbf{a}_m^\ast)$, where $\mathrm{BERT}(\cdot,\cdot)$ denotes the BERT score function, and $\mathbf{a}_m$ and $\mathbf{a}_m^\ast$ are the output and ground-truth answer texts; otherwise, it is computed by
\begin{equation}
\label{eqn:single_round}
t=\lambda\cdot\mathrm{BERT}(\mathbf{a}_m,\mathbf{a}_m^\ast)+(1-\lambda)\cdot\frac{1}{M}\sum_{m=1}^M\mathrm{IoU}(\mathbf{b}_m,\mathbf{b}_m^\ast),
\end{equation}
where $\mathbf{b}_m$ and $\mathbf{b}_m^\ast$ are the detected and ground-truth bounding boxes for the $m$-th object. $\lambda$ is a hyper-parameter that balances the linguistic and visual scores, which is set to be $0.3$ by default.

For multi-round evaluation, due to the logical relationship between subsequent rounds, if the answer in a former round is incorrect (\textit{e.g.}, having detected an incorrect object), the task in a latter round (\textit{e.g.}, asking about the attribute of the object) is no longer meaningful. To reflect this mechanism, we introduce a set of hyper-parameters named truncation thresholds, $\{\tau_n\}_{n=1}^N$, throughout the entire thread, where $N$ is the number of rounds. For any $n$, if $t_n$ computed by~\eqref{eqn:single_round} is smaller than $\tau_n$ (0.3 by default), we immediately terminate the thread and set all scores in the later rounds to be $0$.
The overall multi-round score is the average of all rounds, \textit{i.e.}, $T=\frac{1}{N}\sum_{n=1}^Nt_n$.

\section{The \model Model}
\label{model}

The overall architecture of the \model model is presented in Figure~\ref{fig:framework}. The input data (image and text) is fed to two branches for visual feature extraction and language-related processing. The text output (answer) is directly produced by the multimodal branch. Moreover, when there is a request for localizing visual objects and/or regions, a separate embedding designed for querying is integrated with the visual features and fed to a standalone visual grounding module. We will now describe each module in detail. We denote the input image as $\mathbf{x}_\mathrm{img}$ and the input text as $\mathbf{x}_\mathrm{txt}$.

\subsection{Individual Modules}
\label{model:modules}

\noindent\textbf{Visual feature extraction.}
We resize $\mathbf{x}_\mathrm{img}$ into $512\times512$ and feed it into an iTPN-B model~\cite{tianIntegrallyPreTrainedTransformer2023} that is pre-trained on Object365~\cite{shao2019objects365}, which takes HiViT~\cite{zhang2022hivit} as backbone. The output is a set of features with resolutions of $128\times128$, $64\times64$, $32\times32$, and $16\times16$, respectively, denoted as $\{\mathbf{f}_\mathrm{img}\}$. As we shall see later, $\{\mathbf{f}_\mathrm{img}\}$ is only used for visual grounding.

\noindent\textbf{Multimodal feature extraction.}
We feed $\mathbf{x}_\mathrm{txt}$ to the language branch of the CLIP-L/14 model~\cite{radford2021learning}, and $\mathbf{x}_\mathrm{img}$ (resized into $224\times224$) into the vision branch of the same CLIP-L/14 model. The outputs are a set of language tokens, denoted as $\mathbf{f}_\mathrm{txt}$, and a set of $16\times16$ vision tokens, denoted as $\mathbf{f}_\mathrm{img}'$. To process referring, we follow GPT4RoI~\cite{zhangGPT4RoIInstructionTuning2023} to insert a special language token \texttt{[BBOX]} as a placeholder. The token embedding is then replaced by the features extracted from the corresponding region, for which the RoIAlign~\cite{he2017mask,zhangGPT4RoIInstructionTuning2023} operation is performed on the same CLIP-L/14 model.

\noindent\textbf{Multimodal understanding.}
A multimodal model is trained. It takes $\mathbf{f}_\mathrm{txt}$ and $\mathbf{f}_\mathrm{img}'$ as input and output two-fold embeddings. The first set is simply decoded into the text answer, denoted as $\mathbf{z}_\mathrm{ans}$. The second set corresponds to the queries of visual grounding, denoted as $\mathbf{q}_\mathrm{gnd}$, which is only produced when the multimodal model detects a request for localization in the question. The multimodal model is inherited from LLaVA~\cite{liuVisualInstructionTuning2023}, and we apply the LoRA algorithm~\cite{hu2021lora} for fine-tuning.

\noindent\textbf{Visual grounding.}
We use $\mathbf{q}_\mathrm{gnd}$ to query the multi-scale feature set $\{\mathbf{f}_\mathrm{img}\}$ for visual grounding. The module follows an enhanced DETR~\cite{carion2020detr} object detector named DINO~\cite{zhang2022dino}. Differently, to facilitate communication between them, we design a two-stage querying mechanism. In the first stage, we perform cross-attention between $\mathbf{q}_\mathrm{gnd}$ and $\{\mathbf{f}_\mathrm{img}\}$ to generate some mixed tokens and propagate them through a few self-attention layers (\textit{a.k.a.} the encoder) followed by a query selection module. In the second stage, $\mathbf{q}_\mathrm{gnd}$ is expanded in dimension and directly added to the queries generated in the first stage (both the label queries and box queries are generated by the DINO encoder) and the obtained queries are then propagated through a few attention layers (\textit{a.k.a.} the decoder) to produce the set of box proposals and eventually the bounding boxes, $\mathcal{B}_\mathrm{gnd}$. Please refer to the supplementary material for details.

\subsection{Data Pre-processing and Organization}
\label{model:data}

We organize the \benchmark dataset with some external multimodal dialogue data into the following groups. Full details are provided in the supplementary material.
\begin{itemize}
\item\textbf{Group A: visual question answering.} It involves the Q\&A pair without locations (\textit{i.e.}, bounding boxes) in both the questions and answers. We pre-process \dataA and \dataB by removing all locations from the texts. we combine the two sets (\dataA, \dataB) with LLaVA-Instruction-150K~\cite{liuVisualInstructionTuning2023}. These data are sampled at a ratio of $3:2:5$.
\item\textbf{Group B: referring expression.} It involves the Q\&A pairs with locations in the question but not in the answer. Besides \dataC, we pre-process \dataA and \dataB by choosing the ones with locations in the question, replacing the descriptions to the locations by `it' or `the/this region', and removing all locations from the answer. Then, we combine the three sets (\dataA, \dataB, \dataC, orderly) with public datasets. These data are sampled at a ratio of $2:3:5:5$.
\item\textbf{Group C: visual grounding.} It involves the Q\&A pairs with locations in the answer. Besides \dataD, we filter \dataA and \dataB by choosing the ones with locations in the answer. Then, we combine the three sets (\dataA, \dataB, \dataD, orderly) with public datasets. These data are sampled at a ratio of $3:1:2:5$.
\end{itemize}

\subsection{Optimization}
\label{model:optimization}

There are two sources of supervision. For the text output, we compute the auto-regressive cross-entropy loss between $\mathbf{z}_\mathrm{txt}$ and the ground-truth answer, denoted by $\mathcal{L}_\mathrm{txt}$.
For the grounding output (if present), we compute the localization loss (the same as in DINO) between $\mathcal{B}_\mathrm{gnd}$ and the ground-truth set of bounding boxes, denoted as $\mathcal{L}_\mathrm{gnd}$. 
The overall loss is then written as
\begin{equation}
\mathcal{L}_\mathrm{overall}=\lambda_\mathrm{txt}\cdot\mathcal{L}_\mathrm{txt}+\lambda_\mathrm{gnd}\cdot\mathcal{L}_\mathrm{gnd},
\end{equation}
where $\lambda_\mathrm{txt}$ and $\lambda_\mathrm{gnd}$ are coefficients and both of them are set to $1.0$ by default.

In practice, we find that the visual grounding module is a bit difficult to optimize, so we warm up the training procedure by only using the data in Group C in the early stage. After the grounding loss $\mathcal{L}_\mathrm{gnd}$ becomes small, we add the data in Groups A and B for training.

\subsection{Discussions of the Design Principle}
\label{model:discussions}

The design principle of the \model model is to reflect the idea of decomposition, \textit{i.e.}, the LLM serves as a logic controller to understand user's intention, and the visual understanding and recognition ability is offered by external modules including the feature extractor and the visual grounding architecture. This is related to a few prior works (\textit{e.g.}, ViperGPT~\cite{suris2023vipergpt}, HuggingGPT~\cite{shen2023hugginggpt}, Chameleon~\cite{yu2023scaling}, \textit{etc.}).

Compared to another research line (\textit{e.g.}, Kosmos-2~\cite{pengKosmos2GroundingMultimodal2023}, ~\cite{chenShikraUnleashingMultimodal2023}, ~\cite{youFerretReferGround2023}, \textit{etc.}) that trained a universal tokenizer for vision-language understanding, our methodology claims both advantages and disadvantages. On the one hand, Our model is easily implemented and diagnosed with relatively light computational overhead. On the other hand, extending ChatterBox to a uni-model to support tasks beyond referring and grounding requires heavier engineering efforts.

\begin{table*}[t]
\centering
\small
\setlength{\tabcolsep}{0.12cm}
\caption{A quantitative comparison of the \task metrics (see Section~\ref{benchmark:metric}) between \model (our work) and prior works.}
\vspace{-4pt}
\begin{tabular}{l|ccc|ccc|ccc|c}
\toprule
\multirow{2}{*}{Method} & \multicolumn{3}{c|}{Round \#1} & \multicolumn{3}{c|}{Round \#2} & \multicolumn{3}{c|}{Round \#3} & \multirow{2}{*}{\textit{T}} \\
 & $\mathrm{BERT}(\cdot)$ & $\overline{\mathrm{IoU}}(\cdot,\cdot)$ & $t$ & $\mathrm{BERT}(\cdot)$ & $\overline{\mathrm{IoU}}(\cdot,\cdot)$ & $t$ & $\mathrm{BERT}(\cdot)$ & $\overline{\mathrm{IoU}}(\cdot,\cdot)$ & $t$ & \\
\midrule
LLaVA~\cite{liuVisualInstructionTuning2023} & \textbf{0.9353} & -- & -- & 0.9122 & -- & -- & 0.9002 & -- & -- & -- \\
GPT4RoI~\cite{zhangGPT4RoIInstructionTuning2023} & 0.9157 & -- & -- & 0.8818 & -- & -- & 0.8673 & -- & -- & -- \\
Kosmos-2~\cite{pengKosmos2GroundingMultimodal2023} & 0.9023 & 0.282 & 0.468 & 0.8871 & 0.244 & 0.437 & 0.8712 & 0.137 & 0.357 & 0.421 \\
LISA~\cite{laiLISAReasoningSegmentation2023} & 0.9171 & -- & -- & 0.8822 & -- & -- & 0.8708 & -- & -- & -- \\ 
\midrule
\rowcolor{Gray}
ChatterBox (ours) & 0.9303 & \textbf{0.401 }& \textbf{0.560} & \textbf{0.9184} & \textbf{0.377} & \textbf{0.539} & \textbf{0.9082} & \textbf{0.306} & \textbf{0.487} & \textbf{0.529} \\
\bottomrule
\end{tabular}
\label{tab:mrg}
\end{table*}

\begin{figure*}[!t]
\centering
\includegraphics[width=\linewidth]{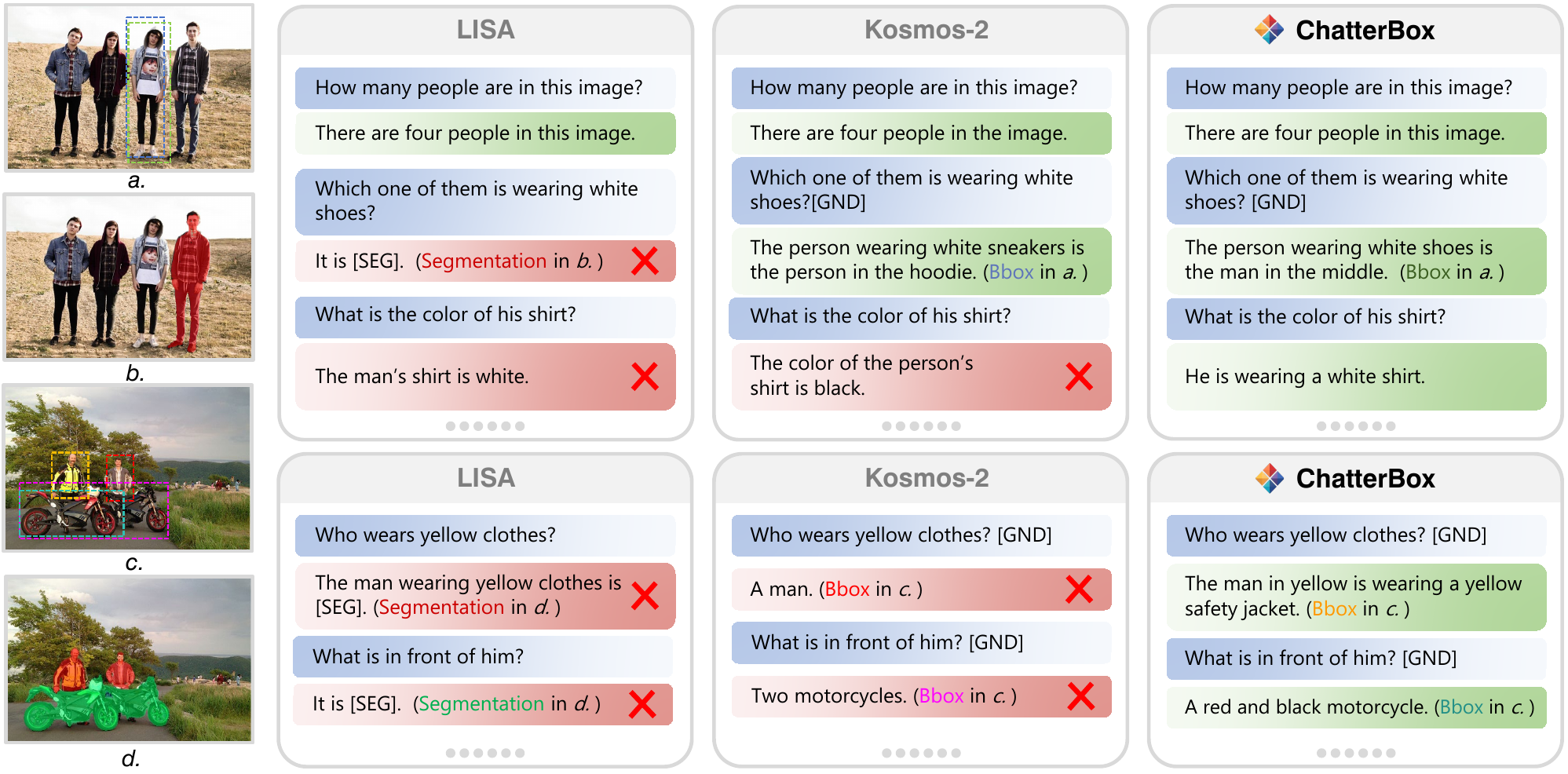}
\caption{A qualitative comparison of multi-round dialogue between LISA~\cite{laiLISAReasoningSegmentation2023}, Kosmos-2~\cite{pengKosmos2GroundingMultimodal2023}, and \model(ours). Our model demonstrates a superior ability to understand multi-round dialogues and perform reasoning (please also refer to the examples in Figure~\ref{fig:task}). We stress that the stronger ability of visual recognition is brought by the explicit vision modules.}
\label{fig:comparison}
\end{figure*}

\section{Experiments}
\label{experiments}

\subsection{Implementation Details}
\label{experiments:details}

\noindent\textbf{Model architecture.}
In the vision branch, we inherit a hierarchical transformer pyramid network (iTPN-B)~\cite{tianIntegrallyPreTrainedTransformer2023} pre-trained on the Objects365 datasets~\cite{shao2019objects365} as the visual encoder, and the DINO detector~\cite{zhang2022dino} as the location decoder which incorporates $300$ queries by default. DINO itself includes an encoder-decoder architecture with $6$ blocks for each part. In the language (multimodal) branch, we use a LLaVA-13B model~\cite{liuVisualInstructionTuning2023}, an MLLM based on LLaMA~\cite{touvron2023llama} and tuned on visual instruction corpus. To fuse the visual features with the query token produced by the LLM, we follow SAM~\cite{kirillov2023segment} to employ a cross-attention operation with a two-way transformer. The individual modules can be replaced by other choices as long as they offer the desired functionality, \textit{e.g.}, vision/language encoding and grounding.

\noindent\textbf{Training configurations.} 
Please refer to Sec.~\ref{model:data} for the details of data preparation and mixture. We utilize $8\times$ NVIDIA A800 GPUs (80GB) for training, making use of DeepSpeed to improve computational efficiency. In the first stage, we employ the AdamW optimizer~\cite{loshchilov2017decoupled} with a learning rate of $0.00005$, zero weight decay, a batch size of $6$, and a gradient accumulation step of $5$. We integrate the WarmupDecayLR learning rate scheduler initialized with a warm-up iteration count of $50$. In the second stage, the learning rate is adjusted to $0.00003$, while the other training parameters remain unchanged. The data from Groups A, B, and C are sampled at a ratio of $2:1:10$, which aims to maximally preserve the ability of visual grounding that we have established in the first stage. The two stages take approximately $1.5$ and $0.5$ days, respectively, and the total training cost is around $15$ GPU-days.

\subsection{Results}

\noindent\textbf{Multi-round Dialogue.}
We first evaluate the entire task, \task, using the metrics defined in Section~\ref{benchmark:metric}. A comparison against prior works is summarized in Table~\ref{tab:mrg}. \textcolor{black}{We curate all threads in the test set of CB-LC into three question-and-answer pairs, where each round (except for the first one) is logically related to the previous rounds, thereby the difficulty increases round by round.}

In terms of the linguistic output, \model produces better $\mathrm{BERT}(\cdot)$ scores than GPT4RoI~\cite{zhangGPT4RoIInstructionTuning2023}, Kosmos-2~\cite{pengKosmos2GroundingMultimodal2023}, and LISA~\cite{laiLISAReasoningSegmentation2023}, and the advantage becomes more significant in the latter two rounds, implying its stronger ability in dealing with multi-round dialogues. \model is slightly inferior to LLaVA~\cite{liuVisualInstructionTuning2023} in the first round but surpasses it in the latter two rounds.

Regarding the visual output, only Kosmos-2 is compared since LLaVA and GPT4RoI cannot perform visual grounding and LISA is unstable in localization\footnote{LISA's lack of support for explicit instructions (\textit{e.g.}, the \texttt{[GND]} token) makes it unable to produce stable localization results. Meanwhile, LISA produces a segmentation mask which sometimes contains outliers that may deteriorate the box-level IoU.}. \textcolor{black}{Similarly, \model achieves the best $\overline{\mathrm{IoU}}(\cdot,\cdot)$ scores throughout the entire thread, and the advantage is even larger than that of the $\mathrm{BERT}(\cdot)$ scores. This is because the grounding quest, calling for the integration of vision and language, is more challenging.} Combining the high quality of linguistic and visual output yields the better \task scores (\textit{i.e.}, $\{t_n\}$ and $T$).

We qualitatively compare \model to Kosmos-2 and LISA (two models equipped with visual grounding) in Figure~\ref{fig:comparison}). Thanks to the specifically collected data for \task and the explicit vision modules, \model shows a stronger ability to accomplish logically complex quests, while the competitors can run into failure. More examples are provided in the supplementary material.

\begin{table}[t]
\centering
\caption{A quantitative comparison of single-round referring expression on the RefCOCOg dataset.}
\vspace{-4pt}
\begin{tabular}{l|c|c|c}
\toprule
Method & METEOR & BERT  \\
\midrule
GPT4RoI~\cite{zhangGPT4RoIInstructionTuning2023} & 9.7 & 0.873 \\
Kosmos-2~\cite{pengKosmos2GroundingMultimodal2023} & 11.5 & 0.871  \\
\rowcolor{Gray}
\midrule
\model & \textbf{14.5} & \textbf{0.880}  \\
\bottomrule
\end{tabular}
\label{tab:experiment_referring}
\end{table}

\begin{table}[t]
\centering
\caption{A quantitative comparison of single-round visual grounding on the COCO~\cite{lin2014microsoft} 2017 test set. Please refer to the main text for the details of prompts and metrics.}
\vspace{-4pt}
\begin{tabular}{l|c|c|c}
\toprule
Method & mIoU & Succ. Rate & mIoU @ Succ. \\
\midrule
Kosmos-2~\cite{pengKosmos2GroundingMultimodal2023} & 0.627 & 0.688 & 0.854 \\
\rowcolor{Gray}
\midrule
\model & \textbf{0.710} & \textbf{0.762} & \textbf{0.904} \\
\bottomrule
\end{tabular}
\label{tab:experiment_grounding}
\end{table}

\noindent\textbf{Single-round referring expression.}
We show that our model trained for \task also enjoys the expected ability of single-round referring expression. We evaluate it on RefCOCOg~\cite{kazemzadeh2014refcoco} and compare it against GPT4RoI and Kosmos-2. Table~\ref{tab:experiment_referring} summarizes the results in terms of the METEOR, CIDER and BERT scores computed upon the captions on given regions. \model shows preferable performance. Evaluation details and some examples are provided in the supplementary material.

\noindent\textbf{Single-round visual grounding.}
Similarly, \model can be used for single-round visual grounding. We compare it against Kosmos-2 on the COCO~\cite{lin2014microsoft} 2017 test set. Table~\ref{tab:experiment_grounding} summarizes the box-level IoU, success rate (IoU is at least $0.5$), and mean IoU of successful cases. Since the MLLMs are sensitive to the prompt, we examine three types of prompts, including (1) `Where is the \texttt{[name]}?', (2) `Can you find the \texttt{[name]}?', and (3) `Can you tell the position of the \texttt{[name]}?', with \texttt{[name]} replaced by the name of object. We report the result of the best prompt for \model and Kosmos-2. As shown, \model surpasses Kosmos-2 in all metrics. Additionally, \model also shows stronger robustness, as the lowest success rate over three prompts is about $0.6$, while the number is around $0.2$ for Kosmos-2. These results are impressive considering that the grounding data is $180\times$ fewer ($500\mathrm{K}$ vs. $90\mathrm{M}$). We owe the ability to the explicit vision module. Some examples are provided in the supplementary material.

\noindent\textbf{Diagnostic studies.}
The first part involves not using the \benchmark data for training. Comparing the first two rows of Table~\ref{tab:diagnosis}, we find that the collected data consistently improves model's ability of \task; similarly, the gain is larger in the second and third rounds. We will release the \benchmark data to facilitate the research in this direction. The second part involves not replacing the concrete object names with pronouns (\textit{e.g.}, `it' or `the object') in the second and third rounds, which degenerates \task into single-round dialogues because the understanding does not rely on the former rounds. Not surprisingly, the model reports similar scores in all three rounds. This indicates that \task indeed increases the difficulty of dialogues so that we believe it is a promising direction for MLLMs.


\begin{table}[!t]
\centering
\setlength{\tabcolsep}{0.03cm}
\caption{Diagnostic results in terms of the BERT score and the $T$ score. \textbf{\benchmark}: whether \benchmark is used for training. \textbf{Ref. Words}: whether pronouns (\textit{e.g.}, `it' or `the object', instead of concrete object names) are used in the inference stage. Note: the third row is \textbf{not} a fair comparison because it is easier than \task.}
\vspace{-4pt}
\begin{tabular}{c|c|c|c|c|c}
\toprule
\benchmark & Ref. Words & Round \#1  & Round \#2 & Round \#3  & \textit{T} \\
\midrule
\cross & \tick  & 0.9291  & 0.9042  & 0.8946   & 0.478\\
\tick  & \tick  & 0.9303  & 0.9184  & 0.9082   & 0.529 \\
\tick  & \cross & 0.9303  & 0.9237  & 0.9210   & 0.547\\
\bottomrule
\end{tabular}
\label{tab:diagnosis}
\end{table}

\section{Conclusions}

We establish a baseline for multi-round multimodal referring and grounding (\task), opening up a promising direction for instance-level multimodal dialogues. In specific, we present a new benchmark and an efficient vision-language model.
The new benchmark, \benchmark, spans challenges including multi-round dialogue, complex spatial relationships among multiple instances, and consistent reasoning, which are beyond those shown in existing benchmarks.
The proposed model, \model, with well-defined feature extraction and optimization strategies, is validated to be very effective in performing multi-round referring and grounding.
With the flexibility to complex instance relationships, the robustness to multiple instances, and the plug-and-play architecture, \model has the potential to significantly advance the multimodal dialogue tasks that involve complicated and precise interactions.

{
    \small
    \bibliographystyle{ieeenat_fullname}
    \bibliography{main}

\begin{thebibliography}{52}
\providecommand{\natexlab}[1]{#1}
\providecommand{\url}[1]{\texttt{#1}}
\expandafter\ifx\csname urlstyle\endcsname\relax
  \providecommand{\doi}[1]{doi: #1}\else
  \providecommand{\doi}{doi: \begingroup \urlstyle{rm}\Url}\fi

\bibitem[Alayrac et~al.(2022)Alayrac, Donahue, Luc, Miech, Barr, Hasson, Lenc, Mensch, Millican, Reynolds, et~al.]{alayrac2022flamingo}
Jean-Baptiste Alayrac, Jeff Donahue, Pauline Luc, Antoine Miech, Iain Barr, Yana Hasson, Karel Lenc, Arthur Mensch, Katherine Millican, Malcolm Reynolds, et~al.
\newblock Flamingo: a visual language model for few-shot learning.
\newblock \emph{Advances in Neural Information Processing Systems}, 35:\penalty0 23716--23736, 2022.

\bibitem[Brown et~al.(2020)Brown, Mann, Ryder, Subbiah, Kaplan, Dhariwal, Neelakantan, Shyam, Sastry, Askell, et~al.]{brown2020language}
Tom Brown, Benjamin Mann, Nick Ryder, Melanie Subbiah, Jared~D Kaplan, Prafulla Dhariwal, Arvind Neelakantan, Pranav Shyam, Girish Sastry, Amanda Askell, et~al.
\newblock Language models are few-shot learners.
\newblock \emph{Advances in neural information processing systems}, 33:\penalty0 1877--1901, 2020.

\bibitem[Carion et~al.(2020)Carion, Massa, Synnaeve, Usunier, Kirillov, and Zagoruyko]{carion2020detr}
Nicolas Carion, Francisco Massa, Gabriel Synnaeve, Nicolas Usunier, Alexander Kirillov, and Sergey Zagoruyko.
\newblock End-to-end object detection with transformers.
\newblock In \emph{European conference on computer vision}, pages 213--229. Springer, 2020.

\bibitem[Chen et~al.(2023{\natexlab{a}})Chen, Qin, Luo, Mi, Li, Sun, and Liu]{chenPositionEnhancedVisualInstruction2023}
Chi Chen, Ruoyu Qin, Fuwen Luo, Xiaoyue Mi, Peng Li, Maosong Sun, and Yang Liu.
\newblock Position-{{Enhanced Visual Instruction Tuning}} for {{Multimodal Large Language Models}}.
\newblock \emph{arXiv preprint arXiv:2308.13437}, 2023{\natexlab{a}}.

\bibitem[Chen et~al.(2023{\natexlab{b}})Chen, Zhang, Zeng, Zhang, Zhu, and Zhao]{chenShikraUnleashingMultimodal2023}
Keqin Chen, Zhao Zhang, Weili Zeng, Richong Zhang, Feng Zhu, and Rui Zhao.
\newblock Shikra: {{Unleashing Multimodal LLM}}'s {{Referential Dialogue Magic}}.
\newblock \emph{arXiv preprint arXiv:2306.15195}, 2023{\natexlab{b}}.

\bibitem[Chiang et~al.(2023)Chiang, Li, Lin, Sheng, Wu, Zhang, Zheng, Zhuang, Zhuang, Gonzalez, et~al.]{chiang2023vicuna}
Wei-Lin Chiang, Zhuohan Li, Zi Lin, Ying Sheng, Zhanghao Wu, Hao Zhang, Lianmin Zheng, Siyuan Zhuang, Yonghao Zhuang, Joseph~E Gonzalez, et~al.
\newblock Vicuna: An open-source chatbot impressing gpt-4 with 90\%* chatgpt quality.
\newblock \emph{See https://vicuna. lmsys. org (accessed 14 April 2023)}, 2023.

\bibitem[Chowdhery et~al.(2022)Chowdhery, Narang, Devlin, Bosma, Mishra, Roberts, Barham, Chung, Sutton, Gehrmann, et~al.]{chowdhery2022palm}
Aakanksha Chowdhery, Sharan Narang, Jacob Devlin, Maarten Bosma, Gaurav Mishra, Adam Roberts, Paul Barham, Hyung~Won Chung, Charles Sutton, Sebastian Gehrmann, et~al.
\newblock Palm: Scaling language modeling with pathways.
\newblock \emph{arXiv preprint arXiv:2204.02311}, 2022.

\bibitem[Chung et~al.(2022)Chung, Hou, Longpre, Zoph, Tay, Fedus, Li, Wang, Dehghani, Brahma, et~al.]{chung2022scaling}
Hyung~Won Chung, Le Hou, Shayne Longpre, Barret Zoph, Yi Tay, William Fedus, Yunxuan Li, Xuezhi Wang, Mostafa Dehghani, Siddhartha Brahma, et~al.
\newblock Scaling instruction-finetuned language models.
\newblock \emph{arXiv preprint arXiv:2210.11416}, 2022.

\bibitem[Dai et~al.(2023)Dai, Li, Li, Tiong, Zhao, Wang, Li, Fung, and Hoi]{instructblip}
Wenliang Dai, Junnan Li, Dongxu Li, Anthony Meng~Huat Tiong, Junqi Zhao, Weisheng Wang, Boyang Li, Pascale Fung, and Steven C.~H. Hoi.
\newblock Instructblip: Towards general-purpose vision-language models with instruction tuning.
\newblock \emph{arXiv preprint arXiv:2305.06500}, 2023.

\bibitem[Devlin et~al.(2018)Devlin, Chang, Lee, and Toutanova]{devlin2018bert}
Jacob Devlin, Ming-Wei Chang, Kenton Lee, and Kristina Toutanova.
\newblock Bert: Pre-training of deep bidirectional transformers for language understanding.
\newblock \emph{arXiv preprint arXiv:1810.04805}, 2018.

\bibitem[Gong et~al.(2023)Gong, Lyu, Zhang, Wang, Zheng, Zhao, Liu, Zhang, Luo, and Chen]{gong2023multimodal}
Tao Gong, Chengqi Lyu, Shilong Zhang, Yudong Wang, Miao Zheng, Qian Zhao, Kuikun Liu, Wenwei Zhang, Ping Luo, and Kai Chen.
\newblock Multimodal-gpt: A vision and language model for dialogue with humans.
\newblock \emph{arXiv preprint arXiv:2305.04790}, 2023.

\bibitem[He et~al.(2017)He, Gkioxari, Doll{\'a}r, and Girshick]{he2017mask}
Kaiming He, Georgia Gkioxari, Piotr Doll{\'a}r, and Ross Girshick.
\newblock Mask r-cnn.
\newblock In \emph{Proceedings of the IEEE international conference on computer vision}, pages 2961--2969, 2017.

\bibitem[Hu et~al.(2021)Hu, Shen, Wallis, Allen-Zhu, Li, Wang, Wang, and Chen]{hu2021lora}
Edward~J Hu, Yelong Shen, Phillip Wallis, Zeyuan Allen-Zhu, Yuanzhi Li, Shean Wang, Lu Wang, and Weizhu Chen.
\newblock Lora: Low-rank adaptation of large language models.
\newblock \emph{arXiv preprint arXiv:2106.09685}, 2021.

\bibitem[Kazemzadeh et~al.(2014)Kazemzadeh, Ordonez, Matten, and Berg]{kazemzadeh2014refcoco}
Sahar Kazemzadeh, Vicente Ordonez, Mark Matten, and Tamara Berg.
\newblock Referitgame: Referring to objects in photographs of natural scenes.
\newblock In \emph{Proceedings of the 2014 conference on empirical methods in natural language processing (EMNLP)}, pages 787--798, 2014.

\bibitem[Kirillov et~al.(2023)Kirillov, Mintun, Ravi, Mao, Rolland, Gustafson, Xiao, Whitehead, Berg, Lo, et~al.]{kirillov2023segment}
Alexander Kirillov, Eric Mintun, Nikhila Ravi, Hanzi Mao, Chloe Rolland, Laura Gustafson, Tete Xiao, Spencer Whitehead, Alexander~C Berg, Wan-Yen Lo, et~al.
\newblock Segment anything.
\newblock \emph{arXiv preprint arXiv:2304.02643}, 2023.

\bibitem[Krishna et~al.(2016)Krishna, Zhu, Groth, Johnson, Hata, Kravitz, Chen, Kalantidis, Li, Shamma, Bernstein, and Li]{krishnaVisualGenomeConnecting2016}
Ranjay Krishna, Yuke Zhu, Oliver Groth, Justin Johnson, Kenji Hata, Joshua Kravitz, Stephanie Chen, Yannis Kalantidis, Li-Jia Li, David~A. Shamma, Michael~S. Bernstein, and Fei-Fei Li.
\newblock Visual {{Genome}}: {{Connecting Language}} and {{Vision Using Crowdsourced Dense Image Annotations}}.
\newblock \emph{arXiv preprint arXiv:1602.07332}, 2016.

\bibitem[Krishna et~al.(2017)Krishna, Zhu, Groth, Johnson, Hata, Kravitz, Chen, Kalantidis, Li, Shamma, et~al.]{krishna2017visual}
Ranjay Krishna, Yuke Zhu, Oliver Groth, Justin Johnson, Kenji Hata, Joshua Kravitz, Stephanie Chen, Yannis Kalantidis, Li-Jia Li, David~A Shamma, et~al.
\newblock Visual genome: Connecting language and vision using crowdsourced dense image annotations.
\newblock \emph{International journal of computer vision}, 123:\penalty0 32--73, 2017.

\bibitem[Lai et~al.(2023)Lai, Tian, Chen, Li, Yuan, Liu, and Jia]{laiLISAReasoningSegmentation2023}
Xin Lai, Zhuotao Tian, Yukang Chen, Yanwei Li, Yuhui Yuan, Shu Liu, and Jiaya Jia.
\newblock {{LISA}}: {{Reasoning Segmentation}} via {{Large Language Model}}.
\newblock \emph{arXiv preprint arXiv:2308.00692}, 2023.

\bibitem[Li et~al.(2022{\natexlab{a}})Li, Li, Xiong, and Hoi]{li2022blip}
Junnan Li, Dongxu Li, Caiming Xiong, and Steven Hoi.
\newblock Blip: Bootstrapping language-image pre-training for unified vision-language understanding and generation.
\newblock In \emph{International Conference on Machine Learning}, pages 12888--12900. PMLR, 2022{\natexlab{a}}.

\bibitem[Li et~al.(2023{\natexlab{a}})Li, Li, Savarese, and Hoi]{liBLIP2BootstrappingLanguageImage2023}
Junnan Li, Dongxu Li, Silvio Savarese, and Steven Hoi.
\newblock {{BLIP-2}}: {{Bootstrapping Language-Image Pre-training}} with {{Frozen Image Encoders}} and {{Large Language Models}}.
\newblock \emph{arXiv preprint arXiv:2301.12597}, 2023{\natexlab{a}}.

\bibitem[Li et~al.(2022{\natexlab{b}})Li, Zhang, Zhang, Yang, Li, Zhong, Wang, Yuan, Zhang, Hwang, Chang, and Gao]{liGroundedLanguageImagePretraining2022a}
Liunian~Harold Li, Pengchuan Zhang, Haotian Zhang, Jianwei Yang, Chunyuan Li, Yiwu Zhong, Lijuan Wang, Lu Yuan, Lei Zhang, Jenq-Neng Hwang, Kai-Wei Chang, and Jianfeng Gao.
\newblock Grounded {{Language-Image Pre-training}}.
\newblock In \emph{2022 {{IEEE}}/{{CVF Conference}} on {{Computer Vision}} and {{Pattern Recognition}}}, pages 10955--10965, 2022{\natexlab{b}}.

\bibitem[Li et~al.(2023{\natexlab{b}})Li, Liu, Wu, Mu, Yang, Gao, Li, and Lee]{li2023gligen}
Yuheng Li, Haotian Liu, Qingyang Wu, Fangzhou Mu, Jianwei Yang, Jianfeng Gao, Chunyuan Li, and Yong~Jae Lee.
\newblock Gligen: Open-set grounded text-to-image generation.
\newblock In \emph{Proceedings of the IEEE/CVF Conference on Computer Vision and Pattern Recognition}, pages 22511--22521, 2023{\natexlab{b}}.

\bibitem[Lin et~al.(2014)Lin, Maire, Belongie, Hays, Perona, Ramanan, Doll{\'a}r, and Zitnick]{lin2014microsoft}
Tsung-Yi Lin, Michael Maire, Serge Belongie, James Hays, Pietro Perona, Deva Ramanan, Piotr Doll{\'a}r, and C~Lawrence Zitnick.
\newblock Microsoft coco: Common objects in context.
\newblock In \emph{Computer Vision--ECCV 2014: 13th European Conference, Zurich, Switzerland, September 6-12, 2014, Proceedings, Part V 13}, pages 740--755. Springer, 2014.

\bibitem[Liu et~al.(2023{\natexlab{a}})Liu, Li, Wu, and Lee]{liuVisualInstructionTuning2023}
Haotian Liu, Chunyuan Li, Qingyang Wu, and Yong~Jae Lee.
\newblock Visual {{Instruction Tuning}}.
\newblock \emph{arXiv preprint arXiv:2304.08485}, 2023{\natexlab{a}}.

\bibitem[Liu et~al.(2023{\natexlab{b}})Liu, Zeng, Ren, Li, Zhang, Yang, Li, Yang, Su, Zhu, et~al.]{liu2023grounding}
Shilong Liu, Zhaoyang Zeng, Tianhe Ren, Feng Li, Hao Zhang, Jie Yang, Chunyuan Li, Jianwei Yang, Hang Su, Jun Zhu, et~al.
\newblock Grounding dino: Marrying dino with grounded pre-training for open-set object detection.
\newblock \emph{arXiv preprint arXiv:2303.05499}, 2023{\natexlab{b}}.

\bibitem[Liu et~al.(2019)Liu, Ott, Goyal, Du, Joshi, Chen, Levy, Lewis, Zettlemoyer, and Stoyanov]{liu2019roberta}
Yinhan Liu, Myle Ott, Naman Goyal, Jingfei Du, Mandar Joshi, Danqi Chen, Omer Levy, Mike Lewis, Luke Zettlemoyer, and Veselin Stoyanov.
\newblock Roberta: A robustly optimized bert pretraining approach.
\newblock \emph{arXiv preprint arXiv:1907.11692}, 2019.

\bibitem[Loshchilov and Hutter(2017)]{loshchilov2017decoupled}
Ilya Loshchilov and Frank Hutter.
\newblock Decoupled weight decay regularization.
\newblock \emph{arXiv preprint arXiv:1711.05101}, 2017.

\bibitem[Mao et~al.(2016)Mao, Huang, Toshev, Camburu, Yuille, and Murphy]{mao2016generation}
Junhua Mao, Jonathan Huang, Alexander Toshev, Oana Camburu, Alan~L Yuille, and Kevin Murphy.
\newblock Generation and comprehension of unambiguous object descriptions.
\newblock In \emph{Proceedings of the IEEE conference on computer vision and pattern recognition}, pages 11--20, 2016.

\bibitem[OpenAI(2023)]{openaiGPT4TechnicalReport2023}
OpenAI.
\newblock {{GPT-4 Technical Report}}.
\newblock \emph{arXiv preprint arXiv:2303.08774}, 2023.

\bibitem[Ouyang et~al.(2022)Ouyang, Wu, Jiang, Almeida, Wainwright, Mishkin, Zhang, Agarwal, Slama, Ray, et~al.]{ouyang2022training}
Long Ouyang, Jeffrey Wu, Xu Jiang, Diogo Almeida, Carroll Wainwright, Pamela Mishkin, Chong Zhang, Sandhini Agarwal, Katarina Slama, Alex Ray, et~al.
\newblock Training language models to follow instructions with human feedback.
\newblock \emph{Advances in Neural Information Processing Systems}, 35:\penalty0 27730--27744, 2022.

\bibitem[Peng et~al.(2023)Peng, Wang, Dong, Hao, Huang, Ma, and Wei]{pengKosmos2GroundingMultimodal2023}
Zhiliang Peng, Wenhui Wang, Li Dong, Yaru Hao, Shaohan Huang, Shuming Ma, and Furu Wei.
\newblock Kosmos-2: {{Grounding Multimodal Large Language Models}} to the {{World}}.
\newblock \emph{arXiv preprint arXiv:2306.14824}, 2023.

\bibitem[Plummer et~al.(2015)Plummer, Wang, Cervantes, Caicedo, Hockenmaier, and Lazebnik]{plummer2015flickr30k}
Bryan~A Plummer, Liwei Wang, Chris~M Cervantes, Juan~C Caicedo, Julia Hockenmaier, and Svetlana Lazebnik.
\newblock Flickr30k entities: Collecting region-to-phrase correspondences for richer image-to-sentence models.
\newblock In \emph{Proceedings of the IEEE international conference on computer vision}, pages 2641--2649, 2015.

\bibitem[Radford et~al.(2021)Radford, Kim, Hallacy, Ramesh, Goh, Agarwal, Sastry, Askell, Mishkin, Clark, et~al.]{radford2021learning}
Alec Radford, Jong~Wook Kim, Chris Hallacy, Aditya Ramesh, Gabriel Goh, Sandhini Agarwal, Girish Sastry, Amanda Askell, Pamela Mishkin, Jack Clark, et~al.
\newblock Learning transferable visual models from natural language supervision.
\newblock In \emph{International conference on machine learning}, pages 8748--8763. PMLR, 2021.

\bibitem[Shao et~al.(2019)Shao, Li, Zhang, Peng, Yu, Zhang, Li, and Sun]{shao2019objects365}
Shuai Shao, Zeming Li, Tianyuan Zhang, Chao Peng, Gang Yu, Xiangyu Zhang, Jing Li, and Jian Sun.
\newblock Objects365: A large-scale, high-quality dataset for object detection.
\newblock In \emph{Proceedings of the IEEE/CVF international conference on computer vision}, pages 8430--8439, 2019.

\bibitem[Shen et~al.(2023)Shen, Song, Tan, Li, Lu, and Zhuang]{shen2023hugginggpt}
Yongliang Shen, Kaitao Song, Xu Tan, Dongsheng Li, Weiming Lu, and Yueting Zhuang.
\newblock Hugginggpt: Solving ai tasks with chatgpt and its friends in huggingface.
\newblock \emph{arXiv preprint arXiv:2303.17580}, 2023.

\bibitem[Sur{\'\i}s et~al.(2023)Sur{\'\i}s, Menon, and Vondrick]{suris2023vipergpt}
D{\'\i}dac Sur{\'\i}s, Sachit Menon, and Carl Vondrick.
\newblock Vipergpt: Visual inference via python execution for reasoning.
\newblock \emph{arXiv preprint arXiv:2303.08128}, 2023.

\bibitem[Thoppilan et~al.(2022)Thoppilan, De~Freitas, Hall, Shazeer, Kulshreshtha, Cheng, Jin, Bos, Baker, Du, et~al.]{thoppilan2022lamda}
Romal Thoppilan, Daniel De~Freitas, Jamie Hall, Noam Shazeer, Apoorv Kulshreshtha, Heng-Tze Cheng, Alicia Jin, Taylor Bos, Leslie Baker, Yu Du, et~al.
\newblock Lamda: Language models for dialog applications.
\newblock \emph{arXiv preprint arXiv:2201.08239}, 2022.

\bibitem[Tian et~al.(2023)Tian, Xie, Wang, Wei, Zhang, Jiao, Wang, Tian, and Ye]{tianIntegrallyPreTrainedTransformer2023}
Yunjie Tian, Lingxi Xie, Zhaozhi Wang, Longhui Wei, Xiaopeng Zhang, Jianbin Jiao, Yaowei Wang, Qi Tian, and Qixiang Ye.
\newblock Integrally {{Pre-Trained Transformer Pyramid Networks}}.
\newblock In \emph{2023 {{IEEE}}/{{CVF Conference}} on {{Computer Vision}} and {{Pattern Recognition}}}, pages 18610--18620. {IEEE}, 2023.

\bibitem[Touvron et~al.(2023)Touvron, Lavril, Izacard, Martinet, Lachaux, Lacroix, Rozi{\`e}re, Goyal, Hambro, Azhar, et~al.]{touvron2023llama}
Hugo Touvron, Thibaut Lavril, Gautier Izacard, Xavier Martinet, Marie-Anne Lachaux, Timoth{\'e}e Lacroix, Baptiste Rozi{\`e}re, Naman Goyal, Eric Hambro, Faisal Azhar, et~al.
\newblock Llama: Open and efficient foundation language models.
\newblock \emph{arXiv preprint arXiv:2302.13971}, 2023.

\bibitem[Wang et~al.(2023{\natexlab{a}})Wang, Chen, Chen, Wu, Zhu, Zeng, Luo, Lu, Zhou, Qiao, et~al.]{wang2023visionllm}
Wenhai Wang, Zhe Chen, Xiaokang Chen, Jiannan Wu, Xizhou Zhu, Gang Zeng, Ping Luo, Tong Lu, Jie Zhou, Yu Qiao, et~al.
\newblock Visionllm: Large language model is also an open-ended decoder for vision-centric tasks.
\newblock \emph{arXiv preprint arXiv:2305.11175}, 2023{\natexlab{a}}.

\bibitem[Wang et~al.(2023{\natexlab{b}})Wang, Lv, Yu, Hong, Qi, Wang, Ji, Yang, Zhao, Song, et~al.]{wang2023cogvlm}
Weihan Wang, Qingsong Lv, Wenmeng Yu, Wenyi Hong, Ji Qi, Yan Wang, Junhui Ji, Zhuoyi Yang, Lei Zhao, Xixuan Song, et~al.
\newblock Cogvlm: Visual expert for pretrained language models.
\newblock \emph{arXiv preprint arXiv:2311.03079}, 2023{\natexlab{b}}.

\bibitem[Xuan et~al.(2023)Xuan, Guo, Yang, and Zhang]{xuanPinkUnveilingPower2023}
Shiyu Xuan, Qingpei Guo, Ming Yang, and Shiliang Zhang.
\newblock Pink: {{Unveiling}} the {{Power}} of {{Referential Comprehension}} for {{Multi-modal LLMs}}.
\newblock \emph{arXiv preprint arXiv:2310.00582}, 2023.

\bibitem[You et~al.(2023)You, Zhang, Gan, Du, Zhang, Wang, Cao, Chang, and Yang]{youFerretReferGround2023}
Haoxuan You, Haotian Zhang, Zhe Gan, Xianzhi Du, Bowen Zhang, Zirui Wang, Liangliang Cao, Shih-Fu Chang, and Yinfei Yang.
\newblock Ferret: {{Refer}} and {{Ground Anything Anywhere}} at {{Any Granularity}}.
\newblock \emph{arXiv preprint arXiv:2310.07704}, 2023.

\bibitem[Yu et~al.(2023)Yu, Shi, Pasunuru, Muller, Golovneva, Wang, Babu, Tang, Karrer, Sheynin, et~al.]{yu2023scaling}
Lili Yu, Bowen Shi, Ramakanth Pasunuru, Benjamin Muller, Olga Golovneva, Tianlu Wang, Arun Babu, Binh Tang, Brian Karrer, Shelly Sheynin, et~al.
\newblock Scaling autoregressive multi-modal models: Pretraining and instruction tuning.
\newblock \emph{arXiv preprint arXiv:2309.02591}, 2023.

\bibitem[Zeng et~al.(2022)Zeng, Liu, Du, Wang, Lai, Ding, Yang, Xu, Zheng, Xia, et~al.]{zeng2022glm}
Aohan Zeng, Xiao Liu, Zhengxiao Du, Zihan Wang, Hanyu Lai, Ming Ding, Zhuoyi Yang, Yifan Xu, Wendi Zheng, Xiao Xia, et~al.
\newblock Glm-130b: An open bilingual pre-trained model.
\newblock \emph{arXiv preprint arXiv:2210.02414}, 2022.

\bibitem[Zhang et~al.(2022{\natexlab{a}})Zhang, Li, Liu, Zhang, Su, Zhu, Ni, and Shum]{zhang2022dino}
Hao Zhang, Feng Li, Shilong Liu, Lei Zhang, Hang Su, Jun Zhu, Lionel~M Ni, and Heung-Yeung Shum.
\newblock Dino: Detr with improved denoising anchor boxes for end-to-end object detection.
\newblock \emph{arXiv preprint arXiv:2203.03605}, 2022{\natexlab{a}}.

\bibitem[Zhang et~al.(2022{\natexlab{b}})Zhang, Roller, Goyal, Artetxe, Chen, Chen, Dewan, Diab, Li, Lin, et~al.]{zhang2022opt}
Susan Zhang, Stephen Roller, Naman Goyal, Mikel Artetxe, Moya Chen, Shuohui Chen, Christopher Dewan, Mona Diab, Xian Li, Xi~Victoria Lin, et~al.
\newblock Opt: Open pre-trained transformer language models.
\newblock \emph{arXiv preprint arXiv:2205.01068}, 2022{\natexlab{b}}.

\bibitem[Zhang et~al.(2023)Zhang, Sun, Chen, Xiao, Shao, Zhang, Chen, and Luo]{zhangGPT4RoIInstructionTuning2023}
Shilong Zhang, Peize Sun, Shoufa Chen, Min Xiao, Wenqi Shao, Wenwei Zhang, Kai Chen, and Ping Luo.
\newblock {{GPT4RoI}}: {{Instruction Tuning Large Language Model}} on {{Region-of-Interest}}.
\newblock \emph{arXiv preprint arXiv:2307.03601}, 2023.

\bibitem[Zhang et~al.(2019)Zhang, Kishore, Wu, Weinberger, and Artzi]{zhang2019bertscore}
Tianyi Zhang, Varsha Kishore, Felix Wu, Kilian~Q Weinberger, and Yoav Artzi.
\newblock Bertscore: Evaluating text generation with bert.
\newblock \emph{arXiv preprint arXiv:1904.09675}, 2019.

\bibitem[Zhang et~al.(2022{\natexlab{c}})Zhang, Tian, Xie, Huang, Dai, Ye, and Tian]{zhang2022hivit}
Xiaosong Zhang, Yunjie Tian, Lingxi Xie, Wei Huang, Qi Dai, Qixiang Ye, and Qi Tian.
\newblock Hivit: A simpler and more efficient design of hierarchical vision transformer.
\newblock In \emph{The Eleventh International Conference on Learning Representations}, 2022{\natexlab{c}}.

\bibitem[Zhou et~al.(2020)Zhou, Wang, Liu, Hu, and Zhang]{zhou2020more}
Yuanen Zhou, Meng Wang, Daqing Liu, Zhenzhen Hu, and Hanwang Zhang.
\newblock More grounded image captioning by distilling image-text matching model.
\newblock In \emph{Proceedings of the IEEE/CVF conference on computer vision and pattern recognition}, pages 4777--4786, 2020.

\bibitem[Zhu et~al.(2023)Zhu, Chen, Shen, Li, and Elhoseiny]{zhu2023minigpt}
Deyao Zhu, Jun Chen, Xiaoqian Shen, Xiang Li, and Mohamed Elhoseiny.
\newblock Minigpt-4: Enhancing vision-language understanding with advanced large language models.
\newblock \emph{arXiv preprint arXiv:2304.10592}, 2023.

\end{thebibliography}
}

\appendix

\setcounter{figure}{4}

\section{CB-300K Generation Details}

In this section, we provide a detailed description of the generation procedure of CB-MRG and CB-LC, which are the main parts of CB-300K. We start with introducing the data cleaning procedure.

\subsection{Data Cleaning}

The data cleaning process is mainly implemented on the Visual Genome~\cite{krishnaVisualGenomeConnecting2016} dataset. We primarily encounter two potential situations that may lead to errors. \textbf{Firstly}, some instances in VG have multiple corresponding bounding boxes, making GPT-4 mistakenly interpret them as distinct instances. To mitigate this, we apply NMS to instances with the same name. Additionally, during the NMS process, we assign unique identifiers to boxes that exceed the threshold to avoid ambiguity. \textbf{Secondly}, there are cases in the dataset where a single object has multiple names due to different annotators for the same image. This situation can also lead to errors in the generated dialogues. To address this, we apply NMS to instances with the same name across the entire dataset, discarding instances that exceed a pre-defined threshold.

\subsection{Prompt for CB-MRG}

The prompt for CB-MRG aims to transform object relationships (\textit{e.g.}, a cat~[x11, x12, y11, y12] on a table~[x21, x22, y21, y22]) into dialogues. The main structure of the prompt includes:
\begin{itemize}
    \item \textbf{Input format explanation.} Each object in the provided description has corresponding coordinates. For example: `black and nice motorcycle~[60, 77, 462, 307] on street~[4, 65, 496, 329]'
    \item \textbf{Task definition.} GPT-4 shall propose questions and answer pairs based on the input relationship information. In a dialogue, each following question should be based on the previous answer. Questions can be about object relationships, object actions, object attributes, object status, object types, etc.
    \item \textbf{Question constraints}: Questions can only be proposed when their answer can be verified as correct with the given description.
    \item \textbf{Indexing objects}: While describing multiple relationship details of an image, objects of the same type are differentiated by adding indices in a `name\_number' format.
    \item \textbf{Output format requirements}: Coordinates of objects mentioned in the sentences shall be appended after each question and answer, in the format $<$the name of object1:~[x1, y1, x2, y2], the name of object2:~[x3, y3, x4, y4]$>$. For example:\\
    \textit{Question-1: What is the color of the shirt of the man? $<$man:~[0.1, 0.1, 0.3, 0.5], shirt:~[0.1, 0.2, 0.3, 0.4]$>$}\\
    \textit{Answer-1: The color is red.\\
    $<$shirt:~[0.1, 0.2, 0.3, 0.4]$>$}\\
    \textit{Question-2: Where is the man sitting?\\
    $<$man:~[0.1, 0.1, 0.3, 0.5]$>$}\\
    \textit{Answer-2: The man is sitting on a chair.\\
    $<$chair:[0.1, 0.4, 0.3, 0.6]$>$}
\end{itemize}
The generated Q\&A pairs are not yet ready to be used. On the one hand, the question sometimes gives too strong hints for the answer or still includes requests that are beyond the given information. On the other hand, the answers might lack the corresponding object's location or provide redundant coordinates. These flaws can be harmful to model training and evaluation, so additional instructions are introduced for refinement:
\begin{itemize}
   \item If the question contains information that is part of the answer, remove the redundant description. For example, replace `Where is the white car parking on the street?' with `Where is the white car?'
   \item Only the coordinates of objects explicitly mentioned in the answer, instead of all objects mentioned in the response sentence, shall be provided. For example, Answer: No, the white cup is on the left of the man. $<$cup:~[25, 100, 51, 124]$>$
   \item If multiple objects are part of the answer, list all these objects. Note that certain objects and the answer may have indirect relationships.
   \item If the answer is not related to the image, there is no need to provide coordinates.
   \item Ensure consistency between relationship information and answer statements. For example, given relationship information: `tan dirt~[0, 86, 500, 498] on feet~[189, 457, 304, 492],' if the question is `What is the object on the tan dirt?' the answer should be `I don't know about it since the information is not given,' not `The object on the tan dirt is feet.'
   \item Do not make any location inferences based on coordinate information.
\end{itemize}

\textbf{The full text of the prompt is very long and is provided in} \texttt{prompt\_CB\_MRG.txt}.

\subsection{Prompt for CB-LC}

\textbf{Firstly}, we prompt GPT-4 (see \texttt{prompt\_CB\_LC.txt}) to construct dialogues. This prompt is similar to the one used for CB-MRG but exhibits some differences:
\begin{itemize}
    \item  Relationship chains are defined as `[[object1 relationship1 object2], [object2 relationship2 object]]', \textit{etc.} (e.g., `[papers below full shelf], [full shelf has white books]', \textit{etc.}).
    \item  Each dialogue is based on one relationship chain.
    \item  Questions must follow the sequence of objects in the relationship chain.
    \item  Each question must include the subject (object from the previous answer).
    \item  Here are some examples for GPT-4:\\
    `relationships:~[[a man~[0.1,0.3,0.4,0.7] holding a cup~[0.1,0.4,0.15,0.45]], [a cup~[0.1,0.4,0.15,0.45] has water~[0.1,0.41,0.15,0.45]]];\\
    \textit{Question-1: What is the man holding?\\
    $<$man:~[0.1,0.3,0.4,0.7]$>$}\\
    \textit{Answer-1: The man is holding a cup.\\
    $<$cup:~[0.1,0.4,0.15,0.45]$>$}\\
    \textit{Question-2: What does the cup have?\\
    $<$cup:~[0.1,0.4,0.15,0.45]$>$}\\
    \textit{Answer-2: There is water in the cup.\\
    $<$water:~[0.1,0.41,0.15,0.45]$>$}'
    \item  There are some excluded dialogue examples for GPT-4:\\
    `relationships:~[[a man~[0.1,0.3,0.4,0.7] holding a cup~[0.1,0.4,0.15,0.45]]];\\
    \textit{Question-1: Where is the cup?\\
    $<$cup:~[0.1,0.4,0.15,0.45]$>$}\\
    \textit{Answer-1: The cup is held by the man.\\
    $<$man:~[0.1,0.3,0.4,0.7]$>$}''
\end{itemize}     
     
\textbf{Secondly}, we conduct some manual cleaning operations (see code at \texttt{cb\_lc\_clearning.py}) to filter out the dialogues that violate prompt requirements, for example:
\begin{itemize}
    \item Objects appear in questions or subjects appear in answers.
    \item Objects in the current question do not appear in the previous answer.
    \item Generated dialogues without corresponding object coordinates.
\end{itemize}

\textbf{Finally}, we use GPT-4 to clean up the dialogues, deleting responses that do not meet the requirements of the questions.
\begin{itemize}
    \item GPT-4 shall determine(see \texttt{cb\_lc\_cleaning\_1.txt}) whether the answer addresses the question and delete dialogues where any round fails to meet the prompt requirements.
    \item GPT-4 shall determine(see \texttt{cb\_lc\_cleaning\_2.txt}) whether the dialogue is contradictory to the given description and delete dialogues where any round contradicts the relationship information (mainly checking for position errors).
\end{itemize}

Please note that these steps aim to generate logically consistent multi-round dialogues based on the given relationship information while ensuring compliance with prompt requirements. Each step involves manual and automated cleaning to refine the generated dialogues.



\section{Experimental Details}
In this section, we provide additional clarification on the LLM query on the DINO detector and the used public datasets.

\begin{figure}[!t]
\centering
\includegraphics[width=0.48\textwidth]{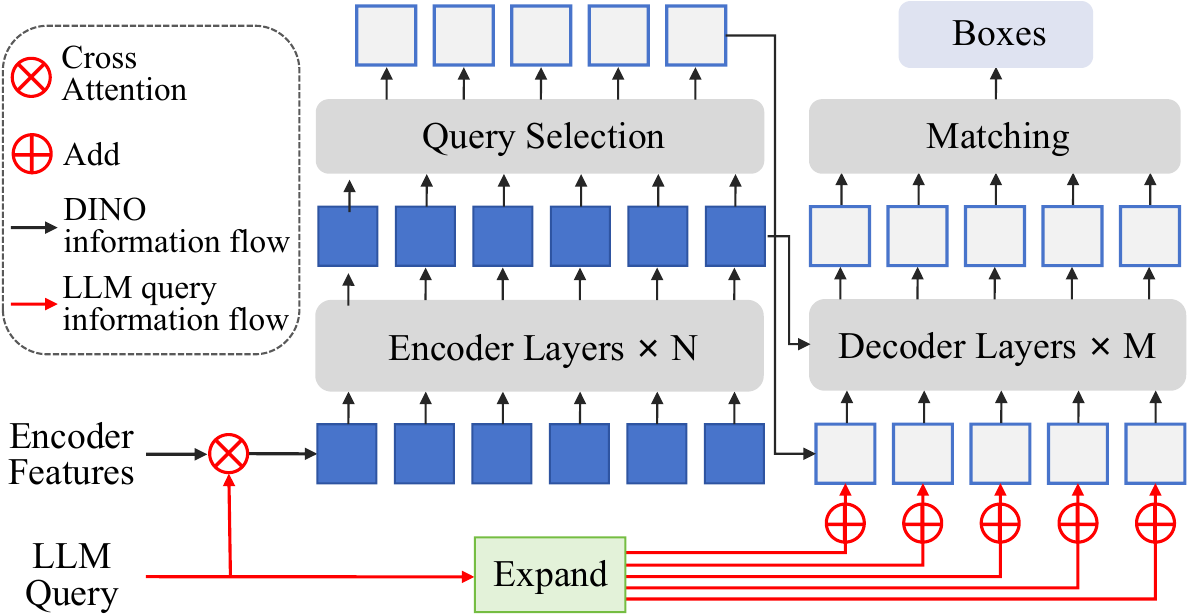}
\caption{The architectural visualization of target-guided query.}
\label{fig:tgt_query}
\end{figure}

\subsection{Target-guided Query}
To facilitate communication between LLM query and vision features, we design a two-stage querying mechanism (named target-guided query, or the `tgt' query for short). The visualization of the target-guided query is shown in Figure~\ref{fig:tgt_query}.
To transfer the LLM query to the visual detector, we propose the target-guided query to enable the ChatterBox to control the visual detector to predict boxes. The target-guided query contains a two-stage querying mechanism. First, we use a cross-attention operation between the query token of LLM and the visual features of the visual encoder, the new visual features are formed with LLM information for subsequent modules. Second, the query token of the LLM is first expanded into two queries with dimensions of ($bs \times N_\mathrm{queries} \times D$) (where $N_\mathrm{queries}$ denotes the query number of DINO detector, and $D$ is the query dimension), one directly added onto the label queries of the DINO detector, and the other onto the box queries. 

\subsection{Datasets}

We organize the \benchmark dataset with some modified public image-text datasets into the following groups.
\begin{itemize}
\item\textbf{Group A: visual question answering.} It involves the Q\&A pair without locations (\textit{i.e.}, bounding boxes) in both the question-and-answer. We pre-process \dataA and \dataB by removing all locations from the texts. We combine the two sets (\dataA, \dataB) with LLaVA-Instruction-150K~\cite{liuVisualInstructionTuning2023}. These data are sampled at a ratio of $3:2:5$.
\item\textbf{Group B: referring expression.} It involves the Q\&A pairs with locations in the question but not in the answer. Besides \dataC, we pre-process \dataA and \dataB by choosing the ones with locations in the question, replacing the descriptions to the locations by `it' or `the/this region', and removing all locations from the answer. Then, we combine the three sets (\dataA, \dataB, \dataC, orderly) with external datasets including COCO~\cite{lin2014microsoft}, RefCOCO~\cite{kazemzadeh2014refcoco}, RefCOCO+~\cite{kazemzadeh2014refcoco}, RefCOCOg~\cite{mao2016generation}, Flickr30K~\cite{plummer2015flickr30k}, and Visual Genome~\cite{krishnaVisualGenomeConnecting2016}. These data are sampled at a ratio of $2:3:5:1:1:1:1:1$.
\item\textbf{Group C: visual grounding.} It involves the Q\&A pairs with locations in the answer (the locations may or may not appear in the question). Besides \dataD, we filter \dataA and \dataB by choosing the ones with locations in the answer. Then, we combine the three sets (\dataA, \dataB, \dataD, orderly) with COCO~\cite{lin2014microsoft}, RefCOCO~\cite{kazemzadeh2014refcoco}, RefCOCO+~\cite{kazemzadeh2014refcoco}, and RefCOCOg~\cite{mao2016generation}. These data are sampled at a ratio of $3:1:2:2:1:1:1$.
\end{itemize}
\subsection{Evaluation}
As current MLLMs' answers may contain extra or unrelated parts that, although do no harm to human understanding, interference the evaluation program's process. We conducted a text cleaning work to make MLLMs' answers easier to be evaluated. Words like `it is', `there is', `region0' that do not appear in the evaluation program's ground truth are removed by our cleaning scripts (see \texttt{cb\_eval\_clean.py}).

\section{Visualization Examples}
\begin{figure*}[!t]
\centering
\includegraphics[width=0.85\linewidth]{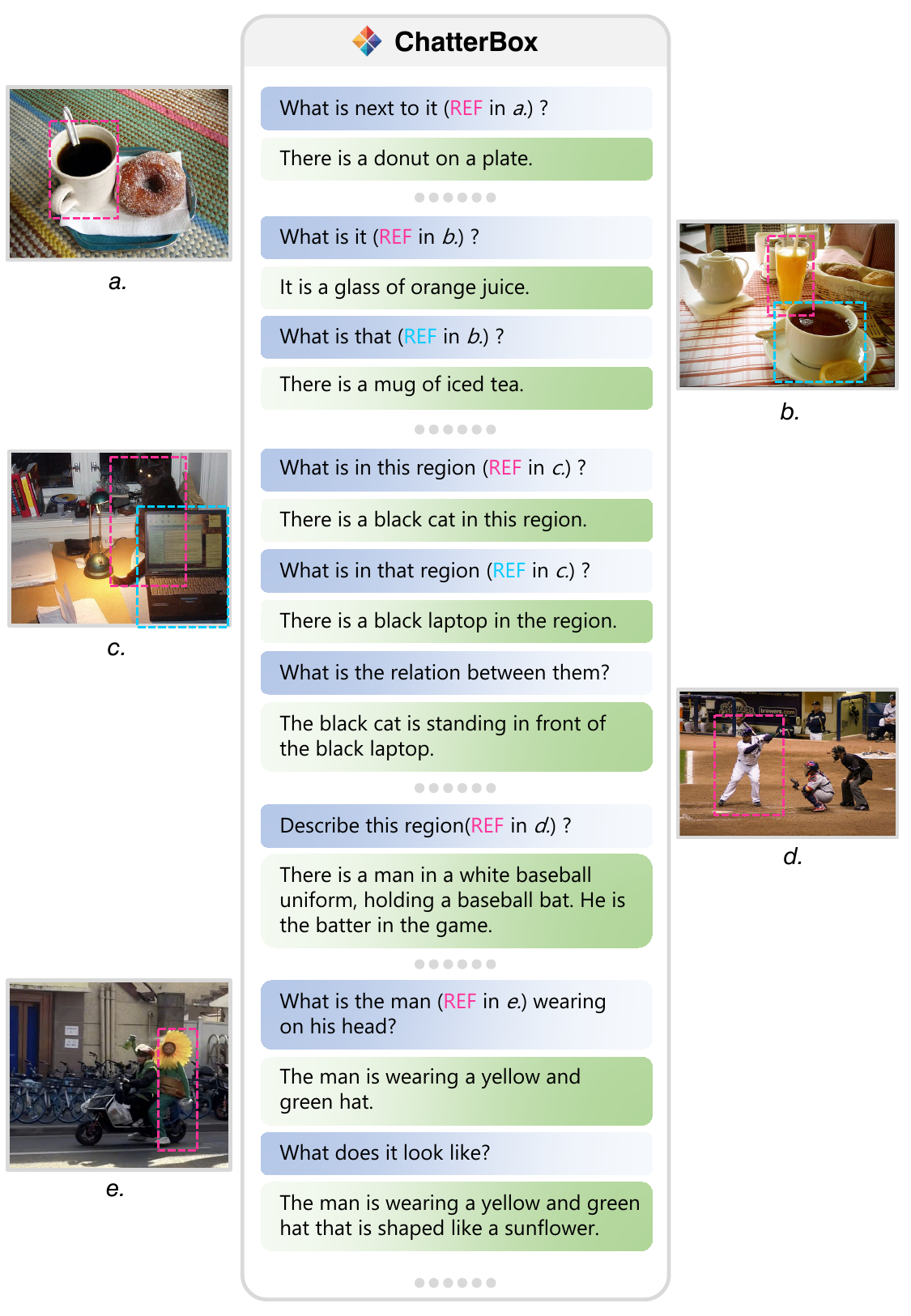}
\caption{Examples of referring expression. \textit{This figure is best viewed in color.}}
\label{fig:supp_vis_referring}
\end{figure*}

\begin{figure*}[!t]
\centering
\includegraphics[width=0.85\linewidth]{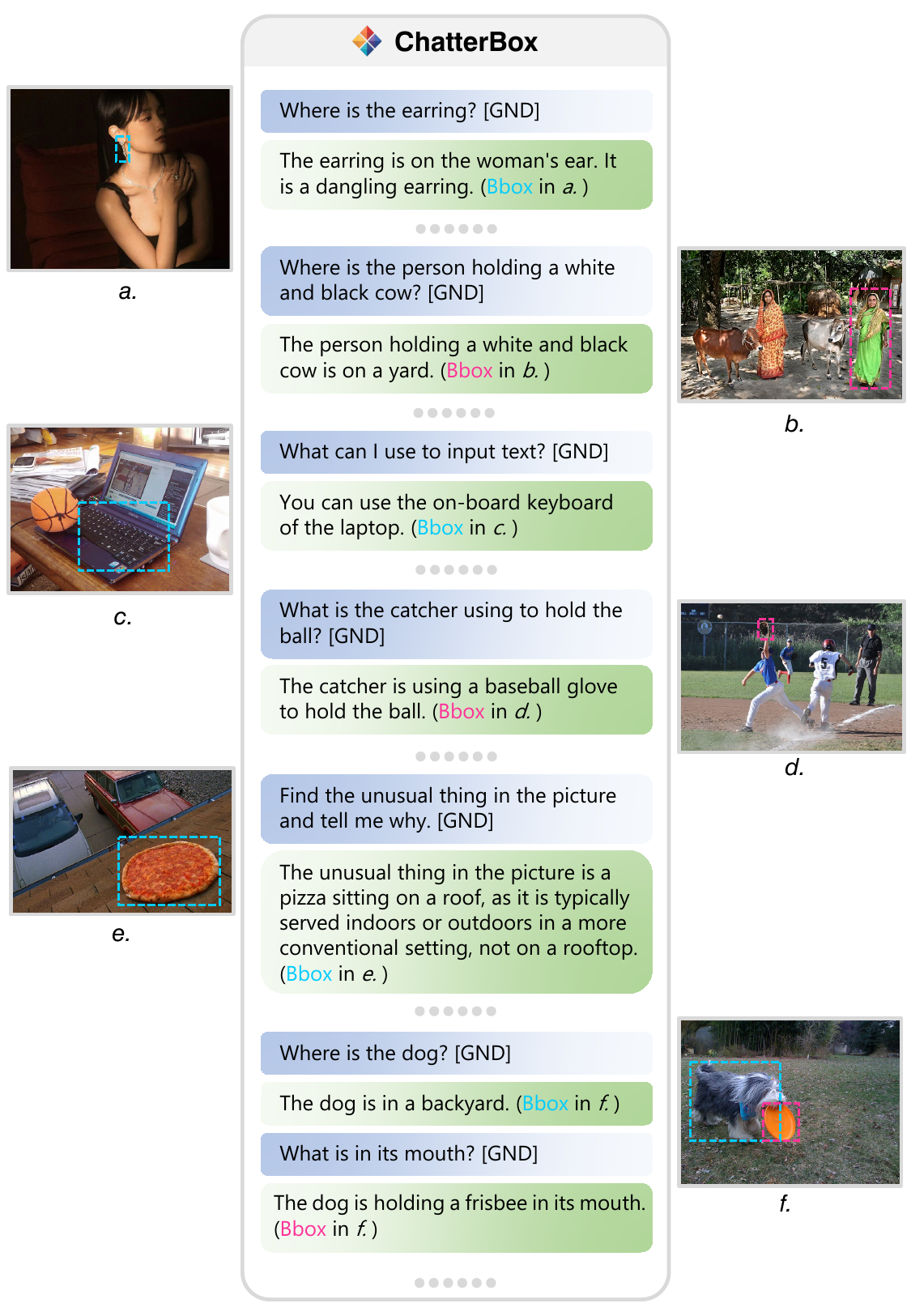}
\caption{Examples of visual grounding. \textit{This figure is best viewed in color.}}
\label{fig:supp_vis_grounding}
\end{figure*}

\begin{figure*}[!t]
\centering
\includegraphics[width=0.85\linewidth]{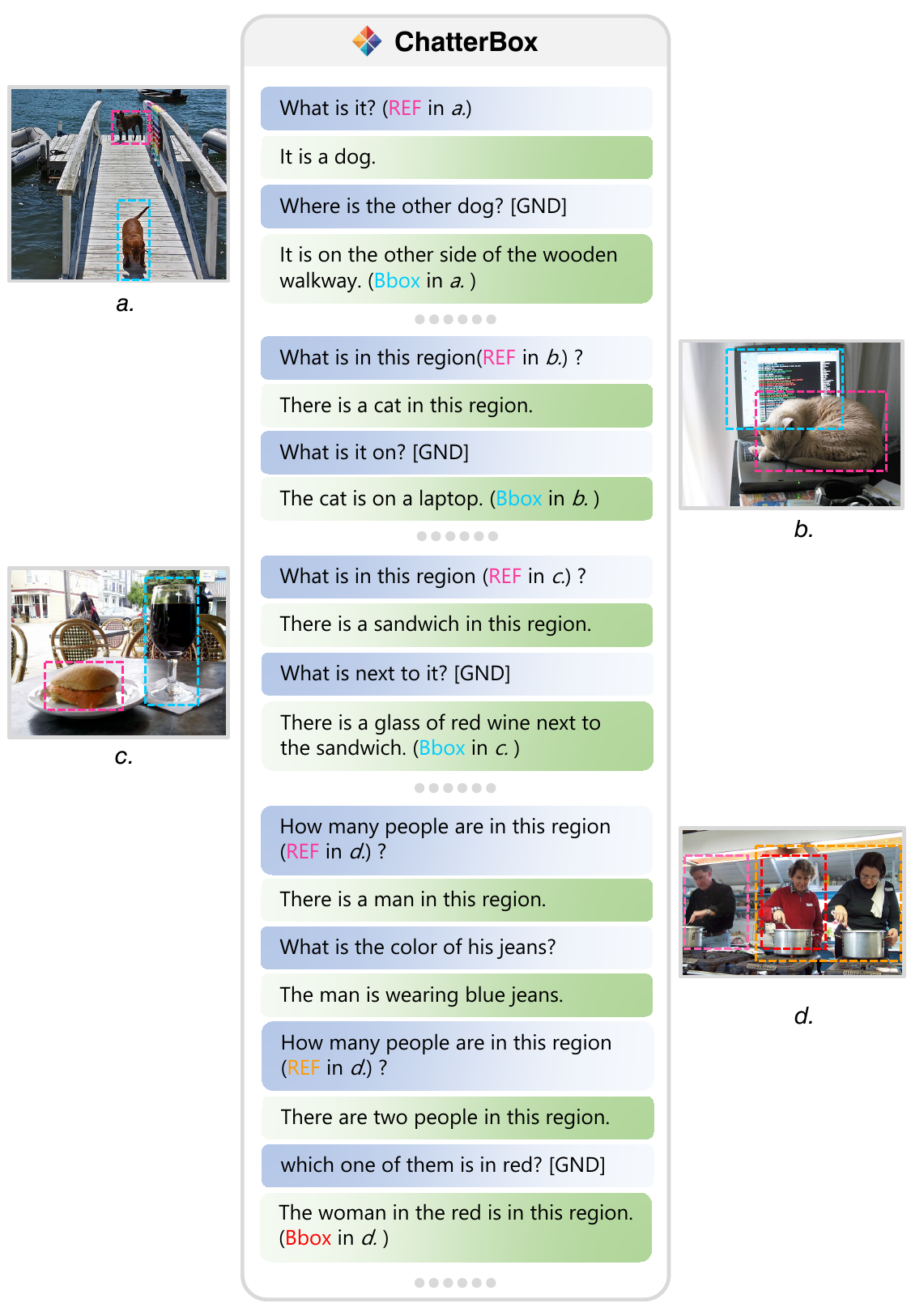}
\caption{Examples of multi-round dialogues. \textit{This figure is best viewed in color.}}
\label{fig:supp_vis_mrg}
\end{figure*}

\begin{figure*}[!t]
\centering
\includegraphics[width=0.85\linewidth]{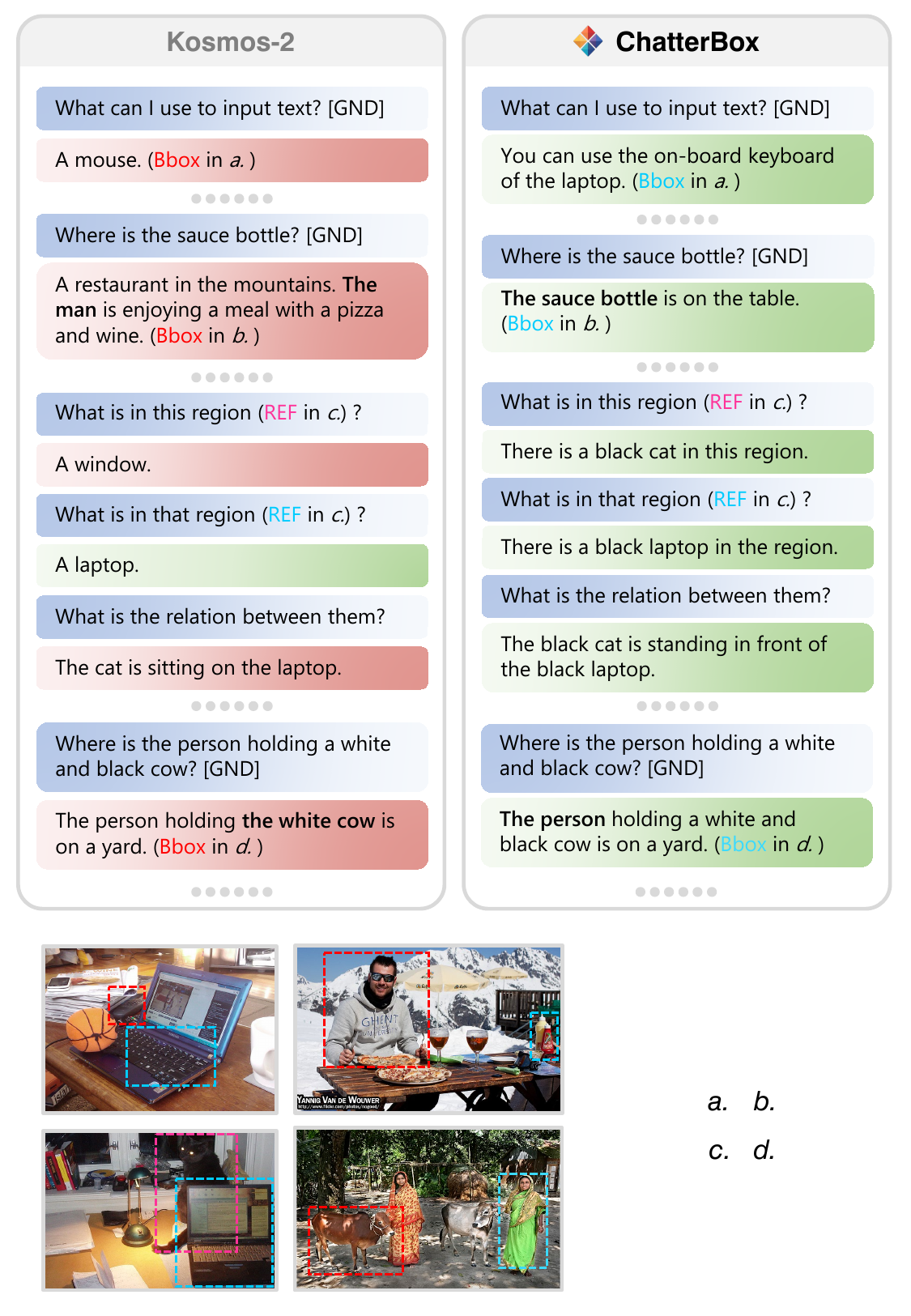}
\caption{The comparison examples with Kosmos-2~\cite{pengKosmos2GroundingMultimodal2023}. \textit{This figure is best viewed in color.}}
\label{fig:supp_vis_comp_with_kosmos2}
\end{figure*}

In this section, we provide more visualization examples. These examples are generated by the ChatterBox agent. We test special ability in terms of referring expression (Figure~\ref{fig:supp_vis_referring}), visual grounding (Figure~\ref{fig:supp_vis_grounding}), multi-round dialogues (Figure~\ref{fig:supp_vis_mrg}), and conduct a comparison with Kosmos-2~\cite{pengKosmos2GroundingMultimodal2023} (Figure~\ref{fig:supp_vis_comp_with_kosmos2}).

\end{document}